\documentclass[nohyperref]{article}

\usepackage{microtype}
\usepackage{graphicx}
\usepackage{subfigure}
\usepackage{booktabs} % for professional tables
\usepackage[ruled,vlined]{algorithm2e}
\usepackage{mathrsfs}
\usepackage{amsmath}
\usepackage{amssymb}

\usepackage{newtxmath}
\usepackage{wrapfig}
\usepackage{amsthm}
\usepackage{thmtools}
\usepackage{thm-restate}
\usepackage{enumitem}
\usepackage[pagebackref,breaklinks,colorlinks]{hyperref}

\mathchardef\mhyphen="2D
\usepackage[accepted]{icml2022}

\usepackage{mathtools}

\usepackage{comment}

\usepackage[capitalize,noabbrev]{cleveref}

\theoremstyle{plain}

\newtheorem{assumption}{Assumption}[section]
\theoremstyle{definition}
\newtheorem{definition}{Definition}[section]

\usepackage[textsize=tiny]{todonotes}

\icmltitlerunning{Matching Learned Causal Effects of Neural Networks with Domain Priors}

\begin{document}

\twocolumn[
\icmltitle{Matching Learned Causal Effects of Neural Networks with Domain Priors}



\icmlsetsymbol{equal}{*}

\begin{icmlauthorlist}
\icmlauthor{Sai Srinivas Kancheti}{equal,iith}
\icmlauthor{Abbavaram Gowtham Reddy}{equal,iith}
\icmlauthor{Vineeth N Balasubramainan}{iith}
\icmlauthor{Amit Sharma}{msr}
\end{icmlauthorlist}

\icmlaffiliation{iith}{Indian Institute of Technology Hyderabad, India}
\icmlaffiliation{msr}{Microsoft Research, Bangalore, India}

\icmlcorrespondingauthor{Sai Srinivas Kancheti}{cs21resch01004@iith.ac.in}

\icmlkeywords{Machine Learning, ICML}

\vskip 0.3in
]



\printAffiliationsAndNotice{\icmlEqualContribution} % otherwise use the standard text.

% neural networks learn patterns from data. they may not be causal because of correlations in the data. this can fail in ood setting
\begin{abstract}
A trained neural network can be interpreted as a structural causal model (SCM) that provides the effect of changing input variables on the model's output. 
However, if training data contains both causal and correlational relationships,
a model that optimizes prediction accuracy 
may not necessarily learn the true causal relationships between input and output variables. On the other hand, expert users often have prior knowledge of the causal relationship between certain input variables and output from domain knowledge. Therefore, we propose a regularization method that aligns the learned causal effects of a neural network with domain priors, including both direct and total causal effects. We show that this approach can generalize to different kinds of domain priors, including monotonicity of causal effect of an input variable on output or zero causal effect of a variable on output for purposes of fairness. Our experiments on twelve benchmark datasets show its utility in regularizing a neural network model to maintain desired causal effects, without compromising on accuracy. Importantly, we also show that a model thus trained is robust and gets improved accuracy on noisy inputs.
\end{abstract}
%Deep Neural Networks (DNNs) have shown great success in many application domains \cite{deng2014deep,Nelson2017StockMP,sadowski2014searching}. 
%The difficulty of understanding how deep neural networks (DNNs) work internally and the need to explain the decisions of DNNs has led to concerted efforts in explainable deep learning \cite{samek_explainable_2019,fan_interpretability_2021,zhang_survey_2020,zhang_visual_2018}. Existing attribution methods for explaining DNN decisions are largely post-hoc (based on an already trained model), and also often based on a learned model function or its gradients. These methods do not inherently capture the \emph{causal} relationships between the input and output variables. 
\vspace{-20pt}
\section{Introduction}
\label{sec:intro}
\vspace{-4pt}
There has been a growing interest in integrating causal principles into machine learning models in recent years. Existing efforts have largely focused on post-hoc explanations of a trained neural network (NN) model's decisions in terms of causal effect \cite{chattopadhyay2019neural,goyal2019explaining}, using counterfactuals for explanations or augmentations  \cite{goyal2019counterfactual,zmigrod2019counterfactual,pitis2020counterfactual,dash2022imagecf}, causal discovery \cite{zhu2019causal}, or embedding causal structures in disentangled representation learning \cite{suter2019robustly,yang2020causalvae}. %, or weighting regularization co-efficients in linear or shallow models \cite{janzing2019causal,bahadori2017causal}. 
However, while there have been efforts of quantifying the causal attributions learned by an NN \cite{chattopadhyay2019neural,goyal2019explaining}, none of these efforts consider the possibility that expert human users that interact with NN models in practice may have prior knowledge of relationships between input and output variables, even causal ones, from domain understanding.
%setting when a user, before training a NN model, may have  knowledge % (complete or partial)of causal relationships between input and output variables. 
In this work, we explore a new methodology -- to the best of our knowledge, the first such effort -- to regularize NN models during training in order to match the \emph{learned causal effects} in NN models with such causal domain priors that are known a priori.

As a simple motivating example, consider the task of predicting the body mass index (BMI) of a person based on features such as miles run each week, calorie intake per day, education, gender, number of dogs one owns, %percentage of fresh vegetables in diet,
and so on. From domain knowledge, an expert user may expect a negative causal effect of miles run per week on BMI (higher the miles, lower the BMI), and a positive causal effect of calorie intake on BMI. %This effect may decrease with increasing number of miles run, following a diminishing returns curve~\cite{kahneman2013choices}. %Percentage of fresh vegetables in diet is correlated with miles run (i.e.,  shares a common cause with miles run, e.g., interest in fitness) and one expects a similar negative effect.  
%They may also expect calorie intake to have a positive causal effect on BMI.
Beyond these, the training data may have unknown, complex correlations. For example, calorie intake could have a complex correlation with miles run---high calorie intake may be correlated both with people who run less as well as with people who run a lot (to support their exercise).
%The above example shows that features may have both \textit{direct} and \textit{indirect} causal effects on the outcome.  
An expert user would expect a trained NN model to learn the causal relationships from input features to BMI while ignoring the correlations in the training data. %, to the extent possible.
%And the direction or shape of these effects may be known to an expert.  
However, given only training data and an accuracy-based loss,  %Without using this domain knowledge, however, 
an NN may rely on correlations to learn a model, whose predictions may not generalize to new data % where the correlations no longer hold or 
and would provide less meaningful feature attributions/explanations to the user\footnote{We use the terms `variables' and `features', as well as 'effects' and 'attributions' interchangeably in this work. %We follow the understanding that causal relationships between input and output variables stay invariant across data sampled from a distribution, as in ~\cite{irm,peters2015causal}.
}. % or provide non-causal explanations. 
E.g., on a training dataset with only high-activity runners who are fit and have a high calorie intake, an NN model may learn that high calorie intake is associated with a lower BMI. While this may be a valid correlation in this training set, it does not reflect the true causal effect between input and output, and will lead to incorrect %(in this case, non-sensical) 
predictions on test data that includes non-runners with high calorie intake. % are present.  may be predicted to have lower BMI. 
Matching the learned causal effects of an NN during training with known domain priors is hence an important goal. %topic of relevance.
%When the shape or direction of the causal effect from feature to output is known, it is hence important to match the feature-to-output relationship learned by a model to the known ground-truth causal relationship. 

Going beyond, a causal effect is understood to be \textit{direct} or \textit{indirect} ~\cite{pearl2009causality}, and it may be important to distinguish between these kind of effects in a trained model. % in a method with the given objective. 
%In addition to distinguishing between correlational and causal effect, it is important to distinguish between the kind of causal effect---\textit{direct} or \textit{indirect}. 
Continuing with our example, while calorie intake and running have a \textit{direct} causal effect on BMI, having pet dogs may enable more miles run or walked in a week which has an \textit{indirect} causal effect on BMI via miles run or walked.
%a feature may cause BMI \textit{indirectly} too. For e.g., 
%So, even though owning a dog does not \textit{directly} affect the BMI, it \textit{indirectly} causes BMI via miles run or walked. 
If the latter is incorrectly construed as direct effect, the model may predict a higher BMI for a person who owns dogs than a person who doesn't own dogs with all other features identical.
%However, a NN model may learn a non-zero \textit{direct} effect of dogs owned by an individual on his/her BMI, which will lead to incorrect predictions at test time (e.g., even with all other identical features, the model may predict a higher BMI for a person who owns dogs than a person who doesn't own dogs). 
Such predictions can reduce the user's trust in the model. Given that causal relationships are known to be stable across different contexts (such as domains with same input-output variables or different datasets sampled from a single population), a model using the correct causal relationships is expected to have better out-of-distribution generalization~\cite{oodsurvey} and be more trustworthy for model-aided decision-making by people~\cite{goodman2018machine}.% and model will likely be discarded. 
\begin{figure}[t]%[11]{r}{0.5\linewidth}
		\centering
        \includegraphics[width=0.83\linewidth]{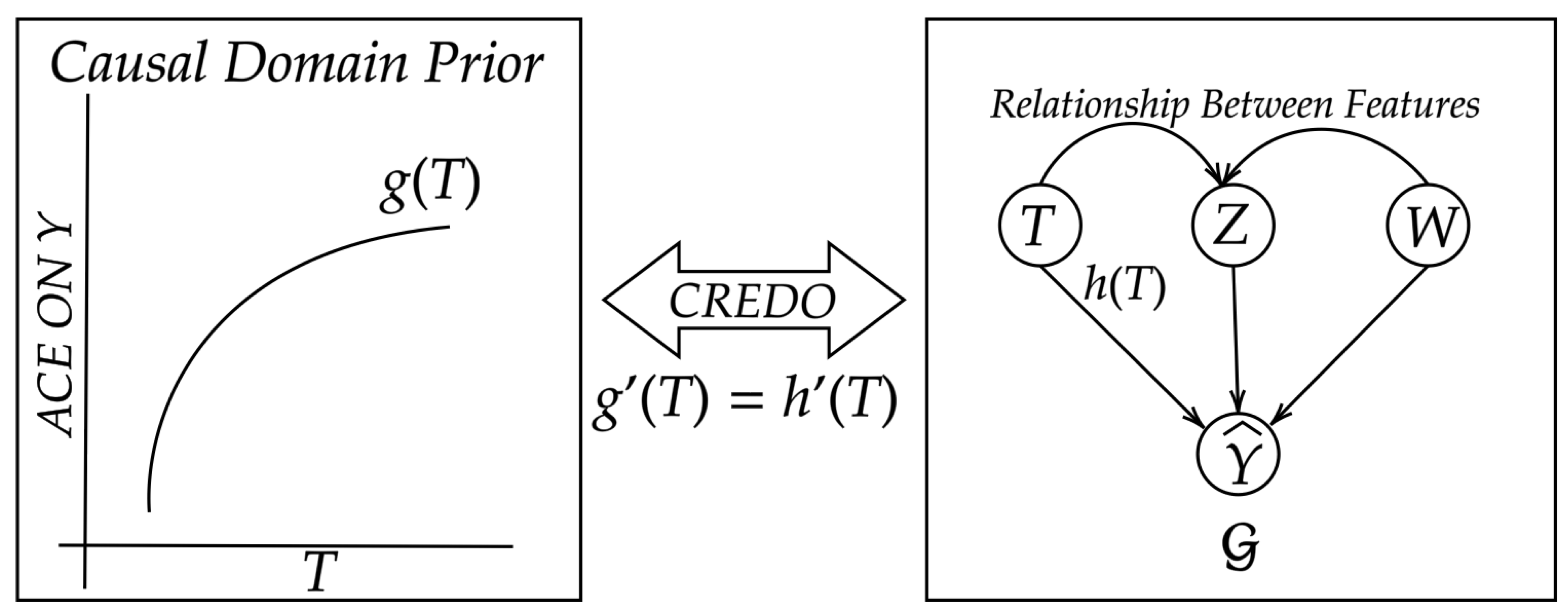}
		\caption{\footnotesize Illustration of our overall idea: CREDO matches the gradient of domain prior function $g(T)$ with the gradient of causal effect $h(T)$ learned by the NN.}
		\label{fig:credo}
        \vspace{-22pt}
\end{figure}

To train NN models consistent with known causal priors, % and thus take a step towards human-machine collaboration in decision-making, 
we need an algorithm to enforce direct and indirect priors during training, while keeping it simple for an expert user to specify the prior.
%\textbf{1)} a simple way for an expert user to specify the causal prior and \textbf{2)} . For \textbf{1)}, 
If not the exact function relating an input variable to output, expert users often know the shape and type of causal effect between some input variables and output. For specific features, they may know whether the effect on outcome is zero, positive or negative, monotonic, following diminishing returns as in economics~\cite{kahneman2013choices}, or following a U- or J-shaped curve as in certain biomedical applications~\cite{salkind-ushaped2010,fraser_nonlinear_2016} (see Secs \ref{sec_methodology} \& \ref{sec_expts} for more examples). We hence work with domain prior functions whose shape and type (direct or otherwise) are provided. We consider \textit{direct} and \textit{total} (combining direct and indirect) effects in this work.
%(direct or total; total effect is the sum of direct and indirect effects. See Sec~\ref{sec_methodology} for details) of the causal effect as a \textit{domain prior}.

To develop our proposed algorithm that matches the learned causal effect of an NN model with domain priors, we define the direct and total \textit{causal effect} of a feature as learned by the NN, and show that the prior can be interpreted as a regularization constraint on the partial derivative (gradient) of the model's output w.r.t. the feature. An illustration of our high-level approach is shown in Fig~\ref{fig:credo}. %Analogous to real-world causal effect, and show how to match it to the provided ground-truth prior. 
Specifically, we propose a gradient-matching regularizer that matches the gradient of the causal effect of the neural network with the gradient of the domain prior during training (see Fig.~\ref{fig:credo}), based on the shape constraints provided by the user.
%The exact gradient of the domain prior is obtained using a heuristic search based on the shape constraints provided by the user.  
Importantly, we validate that the final model learns \textit{causal} effects that are consistent with given domain priors.
The efforts closest to ours are those on enforcing monotonicity or zero attribution of certain input variables on the output  (discussed in Sec \ref{sec:related_work}) such as \cite{sill1998monotonic,sivaraman2020counterexampleguided,ross2017right,rieger2019interpretations}; however, these methods do not consider the \textit{learned causal effect} of the NN model, or generalize to arbitrary shapes relating input variables to output.
%which do not apply to other kinds of domain priors and do not study their models in terms of the causal effects learned. 
Moreover, we study our causal regularizer in the context of both \textit{direct} and \textit{total} causal effects \cite{pearl2009causality}, which is the first such effort that makes such a distinction when understanding causal attributions of input on output in NNs. 
%depending on what information is available in the domain prior. Existing work does not make such distinction between these effects when training a NN model. 
In summary, our key contributions are:
\vspace{-12pt}
\begin{itemize}[leftmargin=*]
\setlength\itemsep{-0.27em}
    \item We introduce and define the notions of direct and total causal effects -- controlled direct, natural direct, and total causal effects, in particular -- learned by an NN model.
    \item We propose a method for \underline{C}ausal \underline{RE}gularization using \underline{D}omain pri\underline{O}rs (CREDO) %to incorporate causal domain priors in neural network models 
    and show formally that an NN model trained using CREDO maintains consistency of the learned causal effect with the given domain prior and type. CREDO is conceptually simple and easy to implement.%both direct and total causal effects with prior knowledge.
    %\item 
    %We analyze our method, and show how our regularizer maintains consistency of causal effect in the learned model w.r.t. the priors.
    %We define and study how CREDO regularizes controlled direct, natural direct, and total causal effects in neural networks. 
    %We also study the connection between regularizability and the availability of causal domain priors.
    %The method supports regularization for all three kinds of effects---controlled direct, natural direct, and total causal effects. We show that estimation of natural direct and total effects requires partial knowledge of the underlying causal graph over input features while controlled direct effect does not. %can be estimated without it. % present details on how NN models can be regularized for these effects when the underlying causal directed acyclic graph (DAG) is known or when it is unknown.
    %When the causal DAG is unknown, we state an impossibility result that we can not regularize total causal effect.
    \item On several real-world and synthetic datasets, our method can be used to enforce different kinds of domain priors, including zero effect, monotonic effect, and other prior shapes. We validate that the causal effects learned by the NN model trained using CREDO is closer to the true prior, while maintaining its test accuracy. We also show that the model trained with CREDO obtains improved accuracy on noisy test data due to the learned causal effects. %with no significant impact on the test accuracy of the model. %. %We analyze the different scenarios under which our causal regularization can be used, and conduct a comprehensive suite of experiments on both synthetic and real-world datasets to show the usefulness of our regularizer. Our results show significant promise in our method. We also show that 
    %the test accuracy does not degrade after applying CREDO.
\end{itemize}
%\amit{A reviewer had commented that it is not implied that accuracy will stay the same. Will need to address it somewhere}
\vspace{-0.3cm}
\section{Related Work}
\label{sec:related_work}
%Regularization can be viewed as a form of Bayesian prior on the model parameters \cite{polson2019bayesian}. However, 
\noindent \textbf{Understanding causal effect learned by an NN.} Our work focuses on maintaining priorly known causal relationships between input and output variables while training an NN model. The first known efforts that attempted to characterize such learned causal effects in an NN model were \cite{alvarez2017causal} and \cite{chattopadhyay2019neural}. While \cite{alvarez2017causal} extended the notion of locally linear approximations in \cite{ribeiro2016should} specifically to language processing tasks and sequence-to-sequence models,  \cite{chattopadhyay2019neural} presented a method derived from first principles of causality for quantifying the Average Causal Effect (ACE) of the input on output variables in any trained NN and a scalable approach to estimate this quantity. Subsequent work such as \cite{khademi2020causal,yadu2021icip,wang2021contrastive} follows the definition of ACE defined therein to quantify the learned causal effect of an NN, which we also leverage in this work.

\vspace{-3pt} \noindent \textbf{Matching causal effect learned by an NN with priors.} One could view the earliest related efforts as  \cite{archer1993application,sill1998monotonic} that aimed to maintain monotonicity in the learnt NN function. \citet{sill1998monotonic} constrained the weights of the first layer (followed by max-min layers) to be positive to ensure the monotonically increasing effect of input on output. This was subsequently extended to partially monotone problems~\cite{daniels2010monotone,dugas2009incorporating} or to Bayesian networks \cite{knowledge_intensive}.
%\cite{daniels2010monotone} extended the above work to the case of partially monotone problems. \cite{dugas2009incorporating} considered a case where the function to be learned is non-decreasing in some arguments and convex in some other. \cite{knowledge_intensive} focused on handling situations specifically in Bayesian networks where qualitative influences can also include synergies or Noisy-Or structures. 
However, these efforts did not consider a causal perspective or study the learned causal effect of the NN model. %the relationships to be causal. % besides, these efforts focused on shallow models, where in case of NNs, it may be sufficient to control the parameters of just the first layer of the NN. \amit{if the definition of shallow models is 1-layer nn, we should specify that}

More recent efforts have attempted to influence the feature attributions of a learned model to be either monotonic~\cite{gupta2015monotonic,you2017deep,gupta2019incorporate,sivaraman2020counterexampleguided,pender} or zero~\cite{ross2017right,rieger2019interpretations}. \citet{rieger2019interpretations} proposed a method to penalize model explanations that did not align with prior knowledge; applicable for enforcing simple constraints like zero feature attribution but not for monotonicity or other prior shapes.
All these efforts do not consider the causal implications. % Besides, the method uses associational attribution strategies and hence, may not conform with the causal effect of input on output. 
Other methods for enforcing a prior depend on matching to an oracle model that outputs the ground-truth prediction for any input. \citet{srinivas2018knowledge} matched the Jacobian of a student network with a teacher network to transfer feature influences from teacher to student.  \citet{sen2018supervising} proposed an active learning algorithm to train using new counterfactual points, generated by changing a single feature and obtaining correct labels from the oracle model. % and then trains the model on the augmented training set, with the goal of matching feature influences w.r.t. the oracle. 
Our key objective of regularizing a model to maintain priors for causal effect is different from these efforts. %In contrast to existing methods, our method's advantage is its generality: our method can not only force a feature's attribution to be zero or monotonic but can also be used to enforce any differentiable function as a causal prior on the feature attribution. %Importantly, we validate our causal regularization method by studying the causal effect of the learned model, which none of the earlier methods consider to the best of our knowledge. 
Moreover, we consider the different types of causal effect in the learned model---controlled direct, natural direct and total~\cite{pearl2001direct}---and identify the different regularizers needed to enforce them. 

%As stated earlier, the aforementioned methods do not view attribution from a causal perspective.
Other efforts with causal implications have  different objectives such as removing confounding features through regularization~\cite{bahadori2017causal,sen2018supervising,janzing2019causal}, or using causal discovery to infer stable features for prediction~\cite{kyono2020castle}.  \citet{bahadori2017causal} used weighted $L_1$ regularization to penalize non-causal attributes in a linear model. % and showed how this can be used to view latent representations in NNs as causal hypotheses. 
\citet{janzing2019causal} related confounding and overfitting in shallow regression models and proposed an $L_1$ or $L_2$ regularizer. % that also learns a co-efficient in $L_1$ or $L_2$ regularization.
The aforementioned methods however do not consider directly regularizing NN models for causal effects or take into account user-provided domain priors.
\vspace{-10pt}
\section{Types of Causal Effects and Domain Priors}%Identifying Causal Effects in Neural Networks}
\begin{wrapfigure}[14]{r}{0.45\linewidth}
\vspace{-7pt}
 		\centering
         \includegraphics[width=0.78\linewidth]{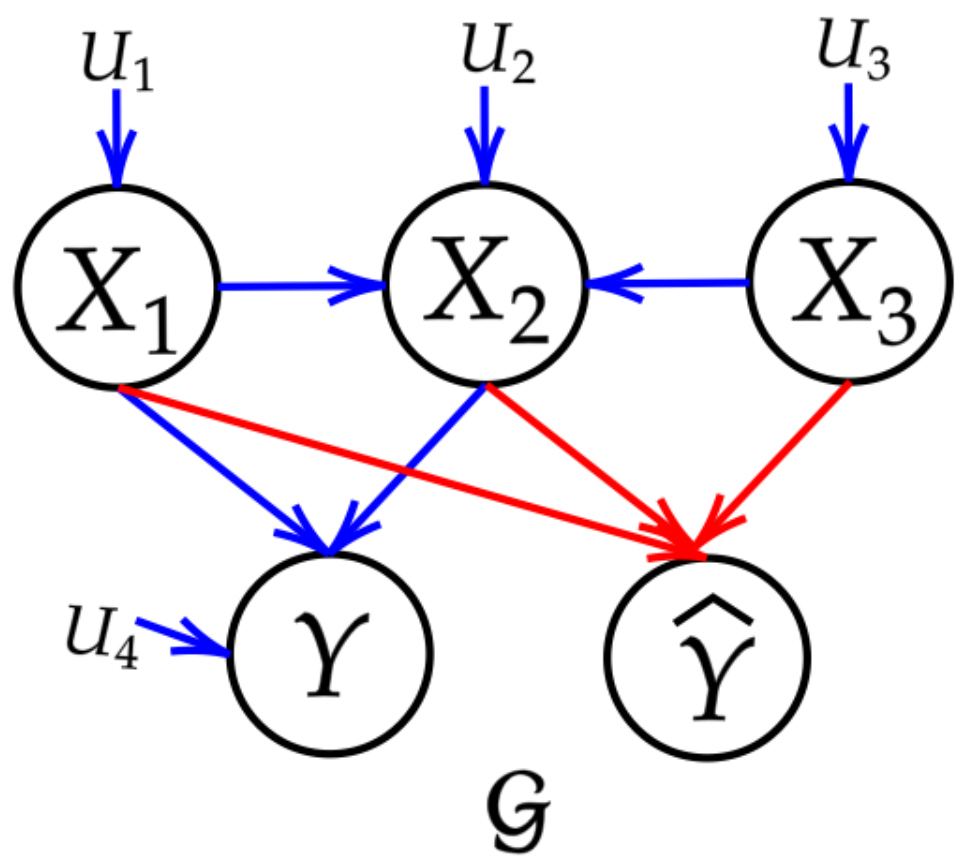}
                  \vspace{-11pt}
 		\caption{Causal graph $\mathcal{G}$ representing input features $X_1,X_2,X_3$, true output $Y$ and NN output $\hat{Y}$ \textit{(Blue arrows = true causal relationships, Red arrows = relationships learned by traditional NN (without CREDO)}.}
 		\label{fig:example}
\end{wrapfigure}

\label{sec_methodology}
\vspace{-4pt}
\noindent \textbf{Problem Setup. } 
Consider an example dataset  with three features $X_1, X_2, X_3$ and outcome $Y$. Its causal graph $\mathcal{G}$ is shown in Fig~\ref{fig:example}. Blue arrows denote the true causal relationship between the variables, while red arrows denote the relationships learned by a traditional feedforward NN model. The graph indicates that only $X_1, X_2$ cause $Y$ in ground truth. In addition, the features are causally connected to each other ($X_1$ and $X_3$ both cause $X_2$). $U_i, i \in \{1,2,3,4\}$,  correspond to the mutually independent error terms for each of these variables. In addition, %a NN $f$ was trained on this dataset. For each input, 
$\hat{Y}=f(X_1, X_2, X_3)$ denotes the prediction of the trained feedforward NN $f$. 
Since neural networks are function approximators~\cite{hornik1989multilayer}, for theoretical analysis, w.l.o.g., we marginalize the hidden layers of the NN and show only input and output nodes, similar to \cite{chattopadhyay2019neural,khademi2020causal}. Note that $X_3$ causes $\hat{Y}$ even though it does not cause the true outcome $Y$, because the NN uses data from all available features. We do not include a noise term for $\hat{Y}$ since it is a deterministic NN function of the inputs.
 
The graph $\mathcal{G}$ helps us formulate key concepts of this work. We denote the \textit{domain prior} as the causal relationship between the features $X_i$ and true outcome $Y$. While the true relationship is often not known, we assume that \textit{properties} of the causal effect from features to $Y$ are known to expert users, e.g., a monotonic effect, U-shaped effect, or zero effect of certain features. For example, a user may know that the direct causal effect of $X_3$ on $Y$ is zero.
In contrast, the relationship learned by the NN corresponds to the edges from features to model prediction $\hat{Y}$ (shown in red in Fig~\ref{fig:example}). In general, there is no guarantee that the relationship $\hat{Y}=f(X_1, X_2, X_3)$ learned by the NN satisfies the properties given for the true causal effect. The goal of this work is to ensure that the learned relationships in NN aligns with the stated properties of true causal effects. 
\vspace{-7pt}
\subsection{Domain Priors}
\label{subsec_domain_priors}
\vspace{-6pt}
%\subsection{Defining Domain Prior \& Learned Causal Effect} 
We define a feedforward NN with inputs $X=\{X_1,X_2,\dots,X_d\}$ as $f:\mathbb{R}^d \rightarrow \{1,2,\dots, C\}$ (for classification) or $f:\mathbb{R}^d \rightarrow \mathbb{R}$ (for regression).  % and output
Analogous to the true causal effect of input features on $Y$ ($\mathbb{E}(Y|do(X))$), we define the causal effect of the features on $\hat{Y}$ implied by the NN, $\mathbb{E}(\hat{Y}| do(X))$, as in \citet{kocaoglu2017causalgan,chattopadhyay2019neural}. For a given feature $X_i$, the ideal goal is to ensure  that $\mathbb{E}(\hat{Y}| do(X_i))=\mathbb{E}(Y|do(X_i))$. In practice, however, a user would not know the true causal effect and instead provide a  domain prior on some \textit{property} satisfied by $\mathbb{E}(Y|do(X))$. Our goal is to ensure that the user-provided property (e.g., monotonicity, $\frac{\partial Y}{\partial do(X_i)} >0$) is also satisfied by the NN's learned  causal effect $\mathbb{E}(\hat{Y}|do(X))$. % (i.e., $\frac{\partial \hat{Y}}{\partial do(X_i)} >0$). 
\vspace{-2pt}  
\begin{definition}
\textbf{Domain Prior.} 
Given a supervised learning dataset $\{({X}^{j}, Y^{j})\}_{j=0}^n$, a domain prior for an input feature $X_i$  is a property satisfied by its true causal effect on $Y$.
\end{definition}
\vspace{-4pt}  
Since the true causal effect (or its properties) cannot be learned from observed data alone, we expect an expert user to provide such priors from domain knowledge. In some cases, the user may also provide the exact \textit{domain prior function}, $g_i^c:\mathbb{R} \rightarrow \mathbb{R}$  from the $i^{th}$ input feature to the ground-truth corresponding to the  $c^{th}$ output neuron (in regression, there is only one output neuron). If not,  given a domain prior shape on $X_i$, we provide a simple  method to estimate the most likely function satisfying the domain prior on a dataset, under some parametric assumptions (see Sec \ref{subsec_credo_algo} and Appendix for details).  %Here $g_i^c$ is a function. % For example, $g_i^c$ may be equal to $2t+3$. While we assume $g_i^c$ to be an explicit function here, in practice, our method can also work if only the shape of $g_i^c$ is available (e.g., monotonic function with a constant positive slope). More information on practical availability of $g_i^c$ is provided at the end of this section.
%We call it the \textit{domain prior function}. 

\vspace{-3pt} \noindent \textbf{Availability of Domain Priors. }
%Our work is based on priors for the causal effect of a feature provided by users as input. In Table~\ref{tab:domain_priors}, we briefly list where such domain priors could come from and what forms they could assume. 
Causal domain priors can come in different forms across fields such as algorithmic fairness, economics, health and physical sciences. % and can be integrated into an NN model during training using our proposed method 
%(see Table~\ref{tab:domain_priors}). 
Specific use cases include: (i) Fairness constraints that require a sensitive attribute's influence on output to be zero -- e.g. skin color in a face classifier~\cite{dash2022imagecf}, or race in a loan approval model~\cite{disc_neurips}; (ii) Monotonic relationship between an input feature and output - e.g., a student's test score may  monotonically affect their admission into a college program, or increasing the number of employees may have a monotonically increasing but diminishing  effect on  a factory's productivity~\cite{kahneman2013choices}; %The nature of these real-world relationships is often a submodular prior~\cite{kahneman2013choices} and can be enforced by our method; %, while prior methods on monotonicity~\cite{archer1993application,sill1998monotonic,daniels2010monotone,gupta2015monotonic,sivaraman2020counterexampleguided} do not provide such guarantees; 
(iii) Arbitrary non-linear functions causally relating input and output - e.g., U-shaped curves have been identified from randomized controlled trials (RCTs) for factors such as cholesterol and diastolic blood pressure~\cite{salkind-ushaped2010} and the dose-response curve for drugs~\cite{calabrese2001u}; or a J-shaped relationship between alcohol and heart disease \cite{fraser_nonlinear_2016}, or  Body Mass Index and mortality~\cite{flanders_adjusting_2008}. A more detailed discussion on priors is included in Appendix~\ref{app:priors}.
% \begin{table}[t]
%     \centering
%     \scalebox{0.7}{
%     \begin{tabular}{lll}
%     \toprule
%          \textbf{Domain} & \textbf{Example Prior} &\textbf{Functional form}\\
%     \midrule
%          Fairness & Zero causal &$\hat{Y}_t=\alpha; \forall \alpha$\\
%          (Dash et al. 2021)(Berk 2019) &effect&\\
%     \midrule
%          Healthcare &U-shape of &$\hat{Y}_t = at^2$\\
%          (Calabrese et al. 2001)(Salkind 2010)&drug effect&\\
%     \midrule
%     Healthcare&J-shape of BMI &$\hat{Y}_t=ae^{bt^2}; x>0$\\
%           (Flanders et al. 2008)(Fraser et al. 2016)  &on mortality&\\
%     \midrule
%          Economics &Diminishing &$\hat{Y}_t=-at^3+bt^2$\\
%         (Kahneman et al. 2013)(Brue 1993) &returns&\\
%     \bottomrule
%     \end{tabular}
%     }
%     \caption{\footnotesize Examples of availability of domain priors ($a,b \in \mathbb{R}^+$).}
%     \label{tab:domain_priors}
% \end{table}
%And there are three corresponding kinds of learned causal effect in NNs, which we define below. 
\vspace{-7pt}
\subsection{Types of Learned Causal Effects in NN}
\label{subsec_causal_effect_types_in_NN}
\vspace{-6pt}
We next define the learned causal effects in an NN model. We consider three kinds of causal effects in this work, as in \cite{pearl2001direct,vanderweele2011controlled}: \emph{controlled direct effect}, \emph{natural direct effect}, and \emph{(natural) total effect}, each of which is defined formally below.
\vspace{-2pt}
\begin{definition}
\textbf{Learned Causal Effect in NN. } Given a feedforward NN $f$, % with inputs $X=\{X_1,X_2,\dots,X_d\}$,  $f:\mathbb{R}^d \rightarrow \{1,2,\dots, C\}$ (for classification) or $f:\mathbb{R}^d \rightarrow \mathbb{R}$ (for regression), % and output $\hat{Y}=f(X)$, 
the learned causal effect of any feature $X_i$ on the NN output $f(X)$ at the end of training is given by $\mathbb{E}(f(X)|do(X_i))$. \end{definition}
\vspace{-4pt}
For convenience, we divide the set of input features $X$ into three disjoint subsets: $T, Z,$ and $W$.
$T$ denotes the feature(s) on which we want to enforce a causal domain prior. $Z=\{Z^1,Z^2,\dots,Z^K\}$ is the set of features that lie on a causal path (in $\mathcal{G}$) between $T$ and $\hat{Y}$ (considering all directed edges, irrespective of color, in Fig~\ref{fig:example}). $W$ denotes the set of remaining features. Aligning with literature in causality~\cite{pearl2009causality}, $T$ is akin to the \textit{treatment} variable,  $\hat{Y}$ is the \textit{target} variable and $Z^1,Z^2,\dots,Z^K$ are the \textit{mediators}. Following~\citet{pearl2009causality}, we use the counterfactual notation $\hat{Y}_t(u)$ to denote the value that $\hat{Y}$ would attain under a specific setting of exogenous noise variables $U=u$ and the intervention $do(T=t)$. 
For the treatment variable $T$, $t^*$ denotes the baseline treatment relative to which causal effects are computed.
Throughout, we make the following assumption on positivity, common in causal inference~\cite{schwab2020learning,shimizu2006linear}.
\vspace{-2pt}
\begin{assumption}
\textbf{Positivity.} $p(T=t|W,Z) >0$ almost surely for all values of $T$, wherever $p(W, Z)>0$.
\end{assumption}
\vspace{-6pt}

\noindent Note that we do not need to assume unconfoundedness since it is always satisfied between $\hat{Y}$ and any feature in $\mathcal{G}$. This is because parents of $\hat{Y}$ are the input features, which are all observed. %ThWe will make a specificWhile we specify Assumption 1 for clarity, our method can also work under certain weaker assumptions, provided in the Appendix.
We start by defining the controlled direct effect.
\begin{definition}
\label{def:controlled_direct_effect}
\textit{(Controlled Direct Effect in NN)}. The Controlled Direct Effect ($NN\mhyphen CDE$) measures the causal effect of treatment $T$ at an intervention $t$ (i.e., $do(T=t)$) on $\hat{Y}$ when all parents of $\hat{Y}$ except $T$ ($Z,W$ in this case) are intervened to pre-defined control values $z,w$ respectively (i.e., $do(Z=z,W=w)$). It is defined as: $NN\mhyphen CDE^{\hat{Y}}_{t}(z, w, u) \coloneqq \hat{Y}_{t,z,w}(u) - \hat{Y}_{t^*,z,w}(u)$. \textit{Average Controlled Direct Effect ($NN\mhyphen ACDE$)} is defined as: $NN\mhyphen ACDE^{\hat{Y}}_{t}(z, w) \coloneqq \mathbb{E}_{U}[\hat{Y}_{t,z,w}]-\mathbb{E}_{U}[\hat{Y}_{t^*,z,w}] =\hat{Y}_{t,z,w}-\hat{Y}_{t^*,z,w}  $.
\end{definition}
\vspace{-3pt}
While the expectation is taken over exogenous noise variables $U$, it can be removed since the neural network $\hat{Y}=f(T,W,Z)$ is a deterministic function which does not depend on the values of the exogenous variables. $t^*$ is a baseline treatment value, as stated earlier (we circumvent the need to specify this value in our method, as discussed in Sec \ref{sec_identify_regularize}). The above definition of $NN\mhyphen ACDE$ is defined for a particular intervention on $\{Z,W\}$ (i.e. all parents of $\hat{Y}$ except $T$) \cite{pearl2001direct}. By our construction, however, the domain priors are expressed only in terms of $T$ and $Y$, so we propose a modified definition for $NN\mhyphen ACDE$ that marginalizes over $\{Z,W\}$. %For regularization, irrespective of the intervention on $\{Z,W\}$, the NN $f$ has to learn a given ACDE of $T=t$ on $\hat{Y}$. 
%So we consider the modified version of ACDE which also
That is, we take the expectation over $\{Z,W\}$ (average of $NN\mhyphen ACDE$ for all interventions on $\{Z,W\}$) along with $U$. Our version of $NN\mhyphen ACDE$ is hence: 
\vspace{-3pt}
\begin{equation*}
\begin{split}
    NN\mhyphen ACDE^{\hat{Y}}_{t} &\coloneqq \mathbb{E}_{Z,W,U}[\hat{Y}_{t,Z,W}]-\mathbb{E}_{Z,W,U}[\hat{Y}_{t^*,Z,W}] \\
    &= \mathbb{E}_{Z,W}[\hat{Y}_{t,Z,W}]-\mathbb{E}_{Z,W}[\hat{Y}_{t^*,Z,W}] 
\end{split}
\end{equation*}
\vspace{-8pt}
\begin{definition}
\label{def:natural}
\textit{(Natural Direct Effect in NN)}. The Natural Direct Effect ($NN\mhyphen NDE$) measures the causal effect of the treatment $T$ at an intervention $t$ (i.e., $do(T=t)$) on $\hat{Y}$ when the nodes of mediating variables $Z$ are intervened to their natural values $Z_{t^*}(u)$ (i.e., $do(Z=Z_{t^*})$) under baseline treatment $do(T=t^*)$. $NN\mhyphen NDE$ is defined as:  $NN \mhyphen NDE^{\hat{Y}}_{t}(u)\coloneqq
\hat{Y}_{t,Z_{t^*}(u)}(u) - \hat{Y}_{t^*,Z_{t^*}(u)}(u)$. \textit{The Average Natural Direct Effect ($NN\mhyphen ANDE$)} is defined as: $NN \mhyphen ANDE^{\hat{Y}}_{t} \coloneqq \mathbb{E}_{U}[\hat{Y}_{t,Z_{t^*}}]-\mathbb{E}_{U}[\hat{Y}_{t^*, Z_{t^*}}]$.
\end{definition}
\begin{definition}

\label{def:total}
\textit{(Total Causal Effect in NN)}. The Total Causal Effect ($NN\mhyphen TCE$) of the treatment $T$ at an intervention $t$ (i.e., $do(T=t)$) on $\hat{Y}$ is given by $NN\mhyphen TCE^{\hat{Y}}_{t}(u) \coloneqq 
\hat{Y}_{t}(u)-\hat{Y}_{t^*}(u)=\hat{Y}_{t, Z_{t}(u)}(u)-\hat{Y}_{t^*, Z_{t^*}(u)}(u)$.
\textit{The Average Total Causal Effect ($NN\mhyphen ATCE$)} is defined as: $NN\mhyphen ATCE^{\hat{Y}}_{t} \coloneqq \mathbb{E}_{U}[\hat{Y}_{t, Z_t}]-\mathbb{E}_{U}[\hat{Y}_{t^*, Z_{t^*}}] = \mathbb{E}_{U}[\hat{Y}_{t}]-\mathbb{E}_{U}[\hat{Y}_{t^*}]$.
%\amit{why is $Z_t$ included in defn of ATCE?}\gautam{I think it was added to be explicit about the values of $Z$. It is redundant, can be removed}
\end{definition}
\vspace{-3pt}
For simplicity, in the rest of the paper, we omit the prefix $NN\mhyphen$ since we always refer to the causal effect in NN when using ACDE, ANDE, ATCE. We use the term Average Causal Effect (ACE) to refer to any of these quantities when the distinction is not necessary.% in a given context.

\vspace{-3pt} \noindent \textbf{Types of domain priors considered. } Since we consider the aforementioned three kinds of causal effects in NN -- \emph{controlled direct}, \emph{natural direct}, and \emph{(natural) total} -- accordingly, there are three kinds of domain priors we consider: controlled direct domain prior, natural direct domain prior, and total domain prior. To illustrate, we continue with the BMI example from Sec~\ref{sec:intro}. Consider a drug that decreases a person's BMI, but which also affects exercise levels. Suppose a randomized trial is conducted to administer the drug and evaluate its effect. Additionally, a free gym membership may be provided to participants of the trial (which would cause people to exercise and thus reduce their BMI). Depending on how the membership was awarded, different kinds of priors may be obtained from the result of the trial. If everyone was provided free gym membership, the result from the trial provides a \textit{controlled direct prior}; the drug's direct effect in the population with gym membership is: $\mathbb{E}[Y_{X=1,M=1}]-\mathbb{E}[Y_{X=0,M=1}]$, where $Y$ is BMI, $X$ is whether drug is given and $M$ is whether gym membership is given ($M=1$ to represent the fact that the individual is performing adequate exercise, due to the gym membership). In contrast, if gym membership was awarded only to people who were already exercising, then the trial result will yield a \textit{natural direct prior}; the effect of the drug at people's base level of exercise: $\mathbb{E}[Y_{X=1,M_{X=0}}]-\mathbb{E}[Y_{X=0,M_{X=0}}]$. Finally, the total effect measures the entire effect of administering the drug, without influencing exercise habits (which may change because of the effects of the drug).
\vspace{-6pt}
\section{Proposed Method: Identification \& Regularization of Causal Effect in NNs}
\label{sec_identify_regularize}
\vspace{-5pt}
We begin by first proving that the causal effects defined above are identifiable and hence can be regularized in NNs. We then provide our algorithm for regularization, along with a simple method to obtain the domain prior function given its shape (or other properties). The proposed regularizer matches learned causal effect in NN to the \textit{domain prior function}, as defined in Sec \ref{subsec_domain_priors}.  

\vspace{-3pt} \noindent \textbf{Matching gradients.}
Given a domain prior function $g_i^c$, our objective is to ensure that the causal effects learned by the NN match the prior. Instead of directly matching causal effects, we rather match the gradient of NN's causal effect with the gradient of $g_i^c$ for three reasons: (i) Properties of prior causal knowledge expressed as a shape (or relative values) is common in many applications, making gradient matching more natural (and avoiding errors in specification of absolute values/constant terms); (ii) There is no closed form expression for the interventional expectation terms (Defns \ref{def:controlled_direct_effect}-\ref{def:total}), thus making gradient matching more computationally efficient; and (iii) gradient matching avoids having to choose a particular baseline treatment value. We hence match the gradient of $g_i^c$ with the gradient of ACE of the $i^{th}$ input feature on $\hat{Y}$ under the graph $\mathcal{G}$ (as in Fig~\ref{fig:credo}).
%For computational efficiency, we match the gradients instead of the exact causal effects since . One more consideration for matching gradients is that in practice, the relative changes in causal effect may be more important in a model than the change in absolute values. If we know by prior knowledge that $g$ changes linearly with $t$, the slope can be chosen as a hyperparameter for which we get the best possible model accuracy, while maintaining the shape of the causal effect. 
%Next, we show how we use gradient matching to regularize for each of the aforementioned causal effects.
For each of the causal effects considered -- controlled direct, natural direct and total, we now show that the effect is identifiable in NNs and then provide regularization procedures.

\vspace{-7pt}
\subsection{Identifying and Regularizing ACDE}
\vspace{-6pt}
Given an NN $f$, a datapoint $(t,z,w)\sim (T,Z,W)$ and its prediction $f(t,z,w)$ (we also use $\hat{Y}(t,z,w)$ to refer to the same quantity), we can intervene on $T$ with  $t'$ and compute $f(t',z,w)$. $f(t',z,w) - f(t,z,w)$ then gives an intuitive expression for the direct effect of $T$ on output. Below we show that this formula captures CDE as in Defn \ref{def:controlled_direct_effect}.
\begin{restatable}[ACDE Identifiability in Neural Networks]{proposition}{acdeidentifiability}
\label{proposition:proposition_cde_identifiability}
 For a neural network with output $\hat{Y}$, the ACDE of a feature $T$ at $t$ on $\hat{Y}$ is identifiable and given by $ACDE^{\hat{Y}}_{t} = \mathbb{E}_{Z,W}\big[\hat{Y}|t,Z,W\big]-\mathbb{E}_{Z,W}\big[\hat{Y}|t^*,Z,W\big]$.
\end{restatable}
\vspace{-3pt}
Proofs of all propositions are in the Appendix for convenience of reading. Since the ACDE is measured as the change in interventional expectation w.r.t. a baseline, the expected gradient of $\hat{Y}$ w.r.t. $t$ at $(t, z, w)$ is equivalent to the gradient of ACDE w.r.t. $t$.
%If we know by prior knowledge that the ACDE of $i^{th}$ feature on $\hat{Y}$ has a form specified by a function $g^{c}_{i}$, we present the following proposition.
\begin{restatable}[ACDE Regularization in Neural Networks]{proposition}{acderegularization}
\label{proposition:proposition_cde}
 The $n^{th}$ partial derivative of $ACDE$ of $T$ at $t$ on $\hat{Y}$ is equal to the expected value of $n^{th}$ partial derivative of $\hat{Y}$ w.r.t. $T$ at $t$, that is: $\frac{\partial^n ACDE^{\hat{Y}}_{t}}{\partial t^n} = \mathbb{E}_{Z,W}\big[\frac{\partial^n [\hat{Y}(t,Z,W)]}{\partial t^n} \big]$.
\end{restatable}
\vspace{-3pt}
Propn \ref{proposition:proposition_cde} allows us to enforce causal priors in an NN model by matching gradients. For an NN model with $d$ inputs and $C$ outputs, let $x^j$ (instance of random variable $X$) denote the $j^{th} d$-dimensional input  to the NN. For a given data point $x^j$, let $\delta G^j$ represent the matrix (of dimension $C\times d$) of derivatives of all available priors $g^i_c$ w.r.t. $x^j$, i.e. $\delta G_{c,i}^j$ denotes the derivative of function $g_i^c$ w.r.t. $i^{th}$ feature of $x^j$. To enforce $f$ to maintain known prior causal knowledge (in terms of gradients), we define our regularizer $R$ as:
\vspace{-9pt}
\begin{equation}
\label{eq:cde_regularizer}
   R(f,G,M) = \frac{1}{N} \sum_{j=1}^{N}\max\{0, \| \nabla_j f \odot M - \delta G^j \|_1-\epsilon \}
\end{equation}
\vspace{-3pt}
where $\nabla_j f$ is the $C \times d$ Jacobian of $f$ w.r.t. $x^j$; $M$ is a $C \times d$ binary matrix that acts as an indicator of features for which prior knowledge is available; $\odot$ represents the element-wise (Hadamard) product; $N$ is the size of training data; and  $\epsilon$ is a hyperparameter to allow a margin of error. For the case where we wish to make gradient of ACDE of a feature to be zero, we set $M_{c,i}=1$ and $\delta G_{c,i}$ to be $0$ and the regularizer hence simplifies into: $R(f,G,M) = \frac{1}{N} \sum_{j=1}^N \max\{0, \| \nabla_j f \odot M \|_1-\epsilon \}$,
which is equivalent to % similar to
the loss function defined in fairness literature \cite{ross2017right,gupta2019incorporate}.
\vspace{-7pt}
\subsection{Identifying and Regularizing ANDE}
\vspace{-6pt}
To identify natural effects in NNs, we need to know: (i) which features belong to the mediating variables set $Z$ (enough to know the partial causal graph containing $Z$);  and (ii) the structural equations of how $Z$ changes when $T$ changes (Defns \ref{def:natural} \& \ref{def:total}). (If we do not know which nodes belong to $Z$, %and the structural equations for $x_i\to Z$  
it is not possible to learn them from training data because there exists a set of causal graphs that are Markov-equivalent w.r.t. a given training distribution, each leading to different causal effects~\cite{pearl2009causality}, making this non-identifiable.) When $Z$ is an empty set, ACDE obtained by controlling for $W$ is equivalent to ANDE~\cite{Zhang2018FairnessID}. %Without any prior information, if we control for all variables $\{Z,W\}$, the causal effect estimator will be biased (e.g., $Z$ should not be controlled while calculating causal effect of $T$ on $\hat{Y}$). Only in the case when $Z$ is an empty set we can control for $\{Z,W\}$ to obtain the unbiased ANDE, which will be ACDE itself~\cite{Zhang2018FairnessID}. 
When $Z$ is known (i.e., mediating variables and their structure) and is non-empty, we now show that we can identify and regularize ANDE in $f$ w.r.t. a baseline treatment value $t^*$, under the below assumption. %Each structural equation of $Z^i$ as a function of its parents can be modeled as separate regressors. In this work, for estimation, we assume that the underlying SCM is linear with Gaussian noise --- this assumption holds true in the datasets we studied.
% prop moved to temp.tex

%To identify ANDE of $T$ where $Z$ are the mediators, we need to make an additional assumption. 
\begin{assumption}\label{ass:unconfound-tz}
(Unconfoundedness of T and Z). $\mathcal{G}$ contains  no unobserved confounders betweeen input features subset $T$ and its mediators $Z$ wrt. $\hat{Y}$, i.e. observed features block all backdoor paths between $T$ and $Z$.
\end{assumption}
\vspace{-3pt}
\begin{restatable}[ANDE Identifiability in Neural Networks]{proposition}{andeidentifiability}
\label{proposition:proposition_nde_identifiability}
Under Assumption~\ref{ass:unconfound-tz}, the ANDE of $T$ at $t$ on $\hat{Y}$ is identifiable and is given by $ANDE^{\hat{Y}}_{t} = \mathbb{E}_{Z_{t^*},W}[\hat{Y}|t,Z_{t^*},W] -\mathbb{E}_{Z_{t^*},W}[\hat{Y}|t^*,Z_{t^*},W]$. % under the assumption that the set $W$ forms a valid adjustment set~\cite{pearl2009causality}.
\end{restatable}
\vspace{-3pt}

\begin{restatable}[ANDE Regularization in Neural Networks]{proposition}{anderegularization}
\label{proposition:proposition_nde}
The $n^{th}$ partial derivative of $ANDE$ of $T$ at $t$ on $\hat{Y}$ is equal to the expected value of $n^{th}$ partial derivative of $\hat{Y}$ w.r.t. $T$ at $t$ , that is, $\frac{\partial^n ANDE^{\hat{Y}}_{t}}{\partial t^n} = \mathbb{E}_{Z_{t^*},W} \big[\frac{\partial^n [\hat{Y}(t,Z_{t^*},W)]}{\partial t^n}\big]$.
\end{restatable}

In this case, R($f$,G,M) is the same as Eqn \ref{eq:cde_regularizer} with the only difference that $\nabla_j f$ is evaluated at $(t,Z_{t^*},W)$.
\vspace{-7pt}
\subsection{Identifying and Regularizing ATCE}
\vspace{-6pt}

Given known $Z$, similar to ANDE, the ATCE of $T$ at $t$ on $\hat{Y}$ is identifiable under assumptions similar to the previous case, as shown below.

\begin{restatable}[ATCE Identifiability in Neural Networks]{proposition}{atceidentifiability}
\label{proposition:proposition_tce_identifiable}
Under Assumption~\ref{ass:unconfound-tz}, the total causal effect of $T$ at $t$ on $\hat{Y}$ is identifiable and is given by $ATCE^{\hat{Y}}_{t} = \mathbb{E}_{Z_t,W}[\hat{Y}|t,Z_{t},W] - \mathbb{E}_{Z_{t^*},W}[\hat{Y}|t^*,Z_{t^*},W]$. % under the assumption that the set $W$ forms a valid adjustment set.
\end{restatable}

\begin{restatable}[ATCE Regularization in Neural Networks]{proposition}{atceregularization}
\label{proposition:proposition_tce}
The gradient of the Average Total Causal Effect (ATCE) of $T$ at $t$ on $\hat{Y}$ is equal to the expected value of the total gradient of $\hat{Y}$ w.r.t. $T$ at $t$, that is, $\frac{d ATCE^{\hat{Y}}_{t}}{dt} = \mathbb{E}_{Z_t,W} \big[\frac{d [\hat{Y}(t,Z_{t},W)}{dt}\big]$.
\end{restatable}

\noindent To regularize the total causal effect, we match the total derivative of $\hat{Y}$ w.r.t. $t$ with the gradient of a given total causal effect prior. %For each $Z^i$, the structural equation $f^i$ can be learned by fitting a regressor on its parents. This is possible since we assume knowledge of the underlying causal structure between input features and we assume that all parents of $Z^i$ are observed. 
The regularizer $R(f,G,M)$ that enforces a NN model to maintain known total causal effect is then:  
\begin{equation}
   R(f,G,M) = \frac{1}{N} \sum_{j=1}^{N} \max\{0, \| \nabla_{j}^t f \odot M - \delta G^j \|_1-\epsilon \}
\end{equation}

\noindent where $\nabla_{j}^t f$ is the $C\times d$ matrix of total derivatives at input $x^j$. The computation of the total derivative is described in Algorithm \ref{algo:regularizer}.
% The total derivative of $\hat{Y}$ w.r.t $T=t$ is given by $\frac{d\hat{Y}}{dt}= \frac{\partial \hat{Y}}{\partial t}+ \sum_{j=1}^{K}\frac{\partial \hat{Y}}{\partial Z^j}\frac{df^j}{dt}$ (proof in Appendix), where $Z^j$ and $f^j$ denote $Z$ and $f$ for the data point $x^j$. 
Other variables are as defined in Eqn.~\ref{eq:cde_regularizer}.
\vspace{-7pt}
\subsection{Final Algorithm: CREDO}
\vspace{-6pt}
\label{subsec_credo_algo}
To summarize, the overall optimization problem with the proposed CREDO regularizer to train the NN is given by:
\begin{equation}
    \arg\min_{\theta} %\mathcal{L}(\mathbf{X}, Y, \theta, G, M) = 
    ERM+\lambda_1 R(f,G,M)% + \lambda_2 ||\theta||_2^2
\end{equation}
\noindent where $\theta$ are parameters of the NN $f$, $ERM$ stands for traditional Empirical Risk Minimizer over the given dataset (based on loss functions such as cross-entropy loss).
%and the last term represents standard $L_2$-regularization.
The regularizer $R(f,G,M)$ is defined differently for each desired causal effect and is summarized in Algorithm \ref{algo:regularizer}.
\begin{algorithm}
\DontPrintSemicolon
\footnotesize
\SetAlgoLined
\KwResult{Regularizers for ACDE, ANDE, ATCE in $f$.}
\textbf{Input:} $\mathcal{D}=\{(x^j, y^j)\}_{j=1}^N$, $y^j \in \{0,1,\dots,C\}$, $x^j \sim X^j$; $\mathbb{Q} = \{i|\exists \ g^c_i \ for\ some\ c\}$; $\mathbb{G} = \{g_i^c|g_i^c$ is prior for $i^{th}$ feature w.r.t. class c\}; $\mathbb{F} = \{f^1,\dots,f^K\}$ is the set of structural equations of the underlying causal model s.t $f^i$ describes $Z^i$; $\epsilon$ is a hyperparameter

\textbf{Initialize:}
 $j = 1, \delta G^j = \vmathbb{0}_{c\times d} \forall j=1,...,N$, $M = \vmathbb{0}_{c\times d}$
 
\While{$j \leq N$}{
\ForEach{$i\in \mathbb{Q}$}{
\ForEach{$g_i^c \in \mathbb{G}$}{
%\tcc{$\theta$ is a dummy variable}
%$\delta G^j[c,i] = \frac{\partial g_i^c(\theta)}{\partial \theta}\rvert_{x_i^j}$; $M[c,i] = 1$
$\delta G^j[c,i] = \nabla g_i^c\big|_{x_i^j}$; $M[c,i] = 1$

\uCase{1: regularizing ACDE}{
$\nabla_j f[c,i] = \frac{\partial \hat{Y}}{\partial x_i}\rvert_{x^j}$
}
\uCase{2: regularizing ANDE}{
\tcc{causal graph is known}
$t=x_i$\\
$\nabla_j f[c,i] = \frac{\partial \hat{Y}}{\partial x_i}\rvert_{(t^j,z_{t^*}^j,w^j)}$
}
\uCase{3: regularizing ATCE}{
\tcc{causal graph is known}
$\nabla_j f[c,i] = \left[\frac{d \hat{Y}}{d x_i} + \sum_{l=1}^{K}\frac{\partial \hat{Y}}{\partial Z^l}\frac{df^l}{dx_i}\right]\rvert_{x^j}$
}
}
}
$j=j+1$
}
return $\frac{1}{N} \sum_{j=1}^N max \{0, \lVert{\nabla_j f \odot M - \delta G^j\rVert}_1 - \epsilon\}$
 \caption{CREDO Regularizer}
 \label{algo:regularizer}
\end{algorithm}
\vspace{-6pt}
\noindent \textbf{Inferring domain prior function. } %of the prior is not in the scope of this work;we assume the user-provided prior to be correct (i.e., representing the true causal mechanisms connecting features to true $Y$). 
When a user provides the domain prior as a shape (not the exact function), we assume a parametric form (see Table~\ref{tab:domain_priors} for examples) for the prior function and select the parameters for which we obtain best validation accuracy. Typically, expert users may be able to specify the search space for a prior (e.g., linear, quadratic in a range); if not, we assume the simplest parametric form satisfying the prior shape. We provide a detailed discussion and present this hyperparameter search  in Appendix~\ref{app:priorsearch}.  
%Going forward, we believe that methods like ours will encourage users to arrange for causal priors in their respective domains.
We note that this work assumes that the prior comes from a domain expert and is hence correct -- validating the prior's correctness may be an independent future direction by itself.
\begin{table}
    \centering
    \scalebox{0.7}{
    \begin{tabular}{lll}
    \toprule
         \textbf{Domain} & \textbf{Example Prior} &\textbf{Functional form}\\
    \midrule
         Fairness & Zero causal &$\hat{Y}_t=\alpha; \forall \alpha$\\
         (Dash et al. 2021)(Berk 2019) &effect&\\
    \midrule
         Healthcare &U-shape of &$\hat{Y}_t = at^2$\\
         (Calabrese et al. 2001)(Salkind 2010)&drug effect&\\
    \midrule
    Healthcare&J-shape of BMI &$\hat{Y}_t=ae^{bt^2}; x>0$\\
          (Flanders et al. 2008)(Fraser et al. 2016)  &on mortality&\\
    \midrule
         Economics &Diminishing &$\hat{Y}_t=-at^3+bt^2$\\
        (Kahneman et al. 2013)(Brue 1993) &returns&\\
    \bottomrule
    \end{tabular}
    }
    \vspace{-3pt}
    \caption{Examples of availability of domain priors ($a,b \in \mathbb{R}^+$).}
    \label{tab:domain_priors}
\end{table}
\vspace{-9pt}
\section{Experiments and Results}
\label{sec_expts}
\vspace{-6pt}
We conducted a comprehensive suite of experiments on real-world and synthetic datasets with different kinds of domain priors to study the proposed method. Our goal was to evaluate whether models trained with CREDO exhibit the correct causal effect as specified by the domain prior. We also show that the models trained by CREDO get better test set performance on noisy input data.

\vspace{-3pt} \noindent \textbf{Datasets and Priors.}
We use two kinds of datasets: \textbf{1)} Four benchmark synthetic datasets from the BNLearn repository~\cite{scutari2014bayesian} where the causal graph is known and all effects can be computed; and \textbf{2)} Eight real-world datasets without knowledge of true causal graph where only $ACDE$ can be computed. BNLearn datasets are Bayesian networks generated from conditional linear Gaussian mechanisms, so the ground-truth domain priors are derived from the generating equations. For the real-world datasets, prior shapes are derived from domain knowledge. %, as discussed in the paragraph below. 
E.g., in the Boston Housing dataset, we use a ground-truth prior that number of rooms should have an increasing monotonic effect on the price of a house. Details of the 12 datasets and known causal graphs are in the Appendix.

%For the experiments on BNLearn datasets, we know the true causal graph and the nature of underlying causal mechanisms: \textit{Conditional Linear Gaussian Bayesian Mechanisms}. In this case, it is valid to fit separate linear regressors for each node using its parents as predictive factors. Regression coefficients then give the gradient of the true prior. For the experiments on other datasets, we refine the model considering the exact slope value of each prior shape as a hyperparameter (see Appendix for algorithm). This way of searching for unknown prior increase the likelihood of recovering true prior (see Appendix for experimental justification).

\vspace{-3pt} \noindent \textbf{Evaluation Metrics.}
Ideally, we would like to compare the error between $E[\hat{Y}|do(X=x)]$ for an NN model and the domain prior for different values of $X$, but there are two challenges. Firstly, the domain prior is not an exact function but a shape provided by user, except when effect is zero. For non-zero effect priors, if the dataset is synthetic, we assume that domain function prior is the true SCM equation for connecting $X$ and $Y$ (CREDO does not have access to the SCM equations). For real-world datasets, we choose the function that provides best accuracy on a held-out validation dataset, using the hyperparameter search procedure described in Appendix. 
Secondly, it is non-trivial to estimate ATCE, $E[\hat{Y}|do(X_i=x)]$ since the identified backdoor estimand, $E[\hat{Y}|do(X_i=x)]=\sum_w E[\hat{Y}|X_i=x,W=w]P(W=w)$  requires a sum over all values of other features;
ACDE and ANDE have similar expressions. Therefore, we use the method from~\citet{chattopadhyay2019neural} to estimate ATCE, ACDE and ANDE for an NN model. It outputs the causal effect at each value of $X_i$, thus yielding an \textit{ACE curve}. A description of this method is in Appendix for completeness.
%it computes the interventional expectation via a second-order Taylor approximation of the NN.
Given a domain prior function and an ACE estimate of a trained NN, we consider the following metrics to compare them: Root Mean Square Error (RMSE), Frechet score and Pearson correlation coefficient. % measured between ACE of the trained model and the given prior. 
We also report test accuracies to confirm that our regularizer maintains test set performance while incorporating causal priors. 

\vspace{-3pt} \noindent \textbf{Baselines.} We compare against vanilla Empirical Risk Minimization (ERM) without our regularizer in all our experiments. In a setting of making ACDE zero, our method becomes the same as  \cite{ross2017right} and one would obtain the same results; hence we do not compare explicitly. In a setting of making ACDE match a monotonic prior, we compare with Point Wise Loss (PWL) \cite{gupta2019incorporate} and Deep Lattice Networks (DLN) \cite{you2017deep}.   In all results,  ``GT'' (prefixed with the variable) refers to the ground-truth prior provided.
%\amit{Can we compare ACDE of PWL/DLN on all synthetic datasets? Now that we are using search over priors, seems fair to compare}

%\noindent \textbf{Estimating domain prior function. } 
%For the experiments on BNLearn datasets, we know the true causal graph and the nature of underlying causal mechanisms: \textit{Conditional Linear Gaussian Bayesian Mechanisms}. In this case, it is valid to fit separate linear regressors for each node using its parents as predictive factors. Regression coefficients then give the gradient of the true prior. For the experiments on other datasets, we refine the model considering the exact slope value of each prior shape as a hyperparameter (see Appendix for algorithm). This way of searching for unknown prior increase the likelihood of recovering true prior (see Appendix for experimental justification).

%\noindent \textbf{Implementation.} 
%We observed that gradients on output logits work better than softmax or logsoftmax activated outputs. A grid search is performed to fix the regularization coefficient $\lambda_1$. We report the specific values chosen for each dataset in the corresponding result. 

Our code is made available for reproducibility. 
%All experiments were conducted on one NVIDIA GeForce 1080Ti. 
Other experimental details including $Z_t$ estimation,  NN architectures and training hyperparameters are in Appendix~\ref{implementation}, \ref{dags} \& \ref{arch}.

%\paragraph{How Do We Choose Domain Priors?}
%For the experiments on SANGIOVESE and MEHRA, we know the true causal graph and the nature of underlying causal mechanisms: \textit{Conditional Linear Gaussian Bayesian Mechanisms}. In this case, it is valid to fit separate linear regressors for each node using its parents as predictive factors. Regression coefficients then approximate the gradient value of the prior. For the experiments on other datasets, based on the domain knowledge, we choose a particular prior shape and then we refine the model considering the exact slope value of each prior shape as a hyperparameter. This way of searching for unknown prior increase the likelihood of recovering true prior (see Appendix for experimental justification).

% \begin{figure}
% \centering
% \includegraphics[width=.5\textwidth,height=20em]{images/Sangiovese/sangiovese_dag.png}
% \caption{Causal Bayesian network that generates SANGIOVESE dataset}
% \label{fig:sangiovese_dag}
% \end{figure}
\vspace{-7pt}
\subsection{Synthetic Data With Linear or Zero Domain Priors}
%\subsection{Setting 1: Causal Graph (Partially) Known}
\noindent \textbf{SANGIOVESE:}
\begin{figure}%[16]{r}{0.6\linewidth}
\centering
\includegraphics[width=0.48\textwidth]{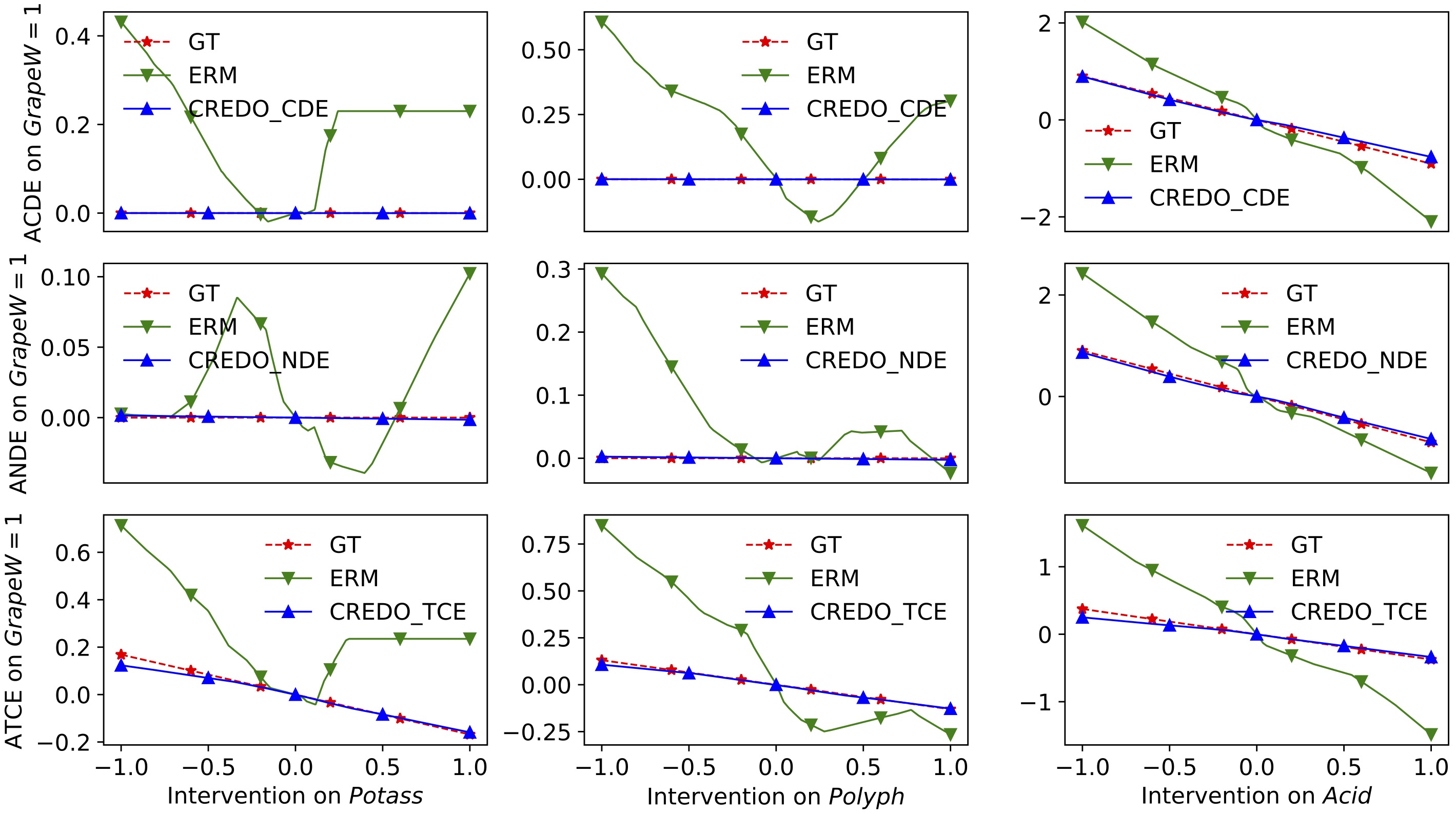}
\vspace{-7pt}
\caption{Results on SANGIOVESE: Comparison of ACDE, ANDE, ATCE learned by ERM and CREDO models.}
\label{fig:sangiovese_plots}
\vspace{-7pt}
\end{figure}
We generate data from the SANGIOVESE conditional linear Gaussian Bayesian network. We consider the output (``GrapesYield'') as a categorical variable and train an NN that predicts if the yield of grapes is better than average. Based on the causal graph and GrapesYield's true structural equation, we choose two kinds of priors: \textbf{1)} \textit{DirectParentPrior:} a linear decreasing ATCE, ANDE and ACDE of \textit{Acid} feature on output (all three effects are the same since Acid is a parent of GrapesYield); and \textbf{2)} \textit{AncestorPrior:} zero ACDE and ANDE for \textit{Potass} (potassium content) and \textit{Polyph} (total polyphenolic content) on output while having a small, non-zero total effect, ATCE. %To obtain structural equations used for regularizing $ANDE$ and $ATCE$, we fit linear regressors for each input variable.
%\amit{From the graph, Potass and Polyph are ancestors of GrapesYield. Why zero total effect in Fig 4?}
Fig \ref{fig:sangiovese_plots}  shows ACE plots and ground-truth ACE for ERM and CREDO. Evidently, CREDO helps conform to the given causal prior for all three features (red and blue lines coincide for ATCE, ANDE and ACDE) whereas ERM model obeys monotonicity only for the \textit{Acid} feature. The quantitative metrics in  Table \ref{tab:sangiovese_mehra} further show the benefit of CREDO regularization. RMSE and Frechet distance are both substantially lower for CREDO than ERM: more than 100 times lower for \textit{Potass} and \textit{Polyph} features, and 16-33\%  lower for \textit{Acid} feature. CREDO matches the prior without a substantial drop in accuracy. Our results on MEHRA, SACHS and Asia datasets from the BNLearn repository showed similar trends, and are reported in the Appendix.
\begin{table}
    \footnotesize
    \centering
    \scalebox{0.85}{
    \centering
    \begin{tabular}{lcccccc}
    \toprule
     Feature & \multicolumn{2}{c}{RMSE} & \multicolumn{2}{c}{Frechet}& \multicolumn{2}{c}{Corr}\\
     \midrule
    &ERM&CREDO&ERM&CREDO&ERM&CREDO\\
    \midrule
    %     \multicolumn{7}{c}{\textbf{SANGIOVESE}}\\
    % \midrule
    \multicolumn{7}{c}{ACDE ($\lambda_1=1.4$) - (Test Acc:ERM: 82.95\%, CREDO: 82.60\%)}\\
     \midrule
     Potass&0.216& \textbf{0.000} &0.431 &\textbf{0.000}&-&-\\
     Polyph&0.273 &\textbf{0.000} &0.607& \textbf{0.000}&-&-\\
     Acid&0.600& \textbf{0.064}& 2.096 &\textbf{0.894} &0.997 &\textbf{0.998}\\
     %Avg. &0.242&\textbf{ 0.020} &0.833&\textbf{ 0.352}& 0.997 &\textbf{0.998}\\
    \midrule
        \multicolumn{7}{c}{ANDE ($\lambda_1=1.0$) - (Test Acc:ERM: 82.95\%, CREDO: 82.75\%)}\\
    \midrule
    Potass&0.043& \textbf{0.000} &0.102 &\textbf{0.001}&-&-\\
     Polyph&0.111&\textbf{0.001}& 0.293&\textbf{0.002}&-&-\\
     Acid&0.684& \textbf{0.046}& 2.425 &\textbf{0.866} &0.993& \textbf{0.999}\\
     %Avg. &0.242&\textbf{0.022}& 0.833& \textbf{0.306}& 0.997 &\textbf{1.000}\\
     \midrule
        \multicolumn{7}{c}{ATCE ($\lambda_1=1.5$) - (Test Acc:ERM: 82.95\%, CREDO: 82.10\%)}\\
    \midrule
    Potass&0.303& \textbf{0.016}& 0.714 &\textbf{0.159}&0.547&\textbf{0.997}\\
     Polyph&0.325&\textbf{0.008} &0.849&\textbf{0.128}&0.937&\textbf{0.998}\\
     Acid&0.423&\textbf{0.025}&1.059&\textbf{ 0.208} &0.547& \textbf{0.996}\\
     %Avg. &0.247& \textbf{0.016} &0.820& \textbf{0.343}& 0.994& \textbf{0.999}\\
    %  \midrule
    % \multicolumn{7}{c}{\textbf{MEHRA}}\\
    % \midrule
    % \multicolumn{7}{c}{ACDE ($\lambda_1=2.2$) -(Test Acc:ERM: 80.75\%, CREDO: \textbf{82.00\%)}}\\
    %   \midrule
    %  Latitude&0.309& \textbf{0.045}& 0.366& \textbf{0.449}& 0.115 &\textbf{0.997}\\
    %  O3&0.588& \textbf{0.011} &0.062 &\textbf{0.991}& -0.494 &\textbf{1.000}\\
    % SO2&0.790&\textbf{ 0.196}& \textbf{3.057} &1.840& 0.983 &\textbf{0.995}\\
    % % Avg. &0.562 &\textbf{0.084}& \textbf{1.162}& 1.093& -0.494&\textbf{0.995}\\
    % \midrule
    %     \multicolumn{7}{c}{ANDE ($\lambda_1=2.1$) - (Test Acc:ERM: 80.75\%, CREDO: 80.05\%)}\\
    % \midrule
    % Latitude&\textbf{0.309}& 0.400& 0.367& \textbf{0.481}& \textbf{0.116} &0.000\\
    %  O3&0.588& \textbf{0.023}& 0.063& \textbf{1.038}& -0.497& \textbf{1.000}\\
    %  SO2& 0.790& \textbf{0.045}& \textbf{3.054} &1.519 &0.983 &\textbf{1.000}\\
    % % Avg. &0.562& \textbf{0.156} &\textbf{1.161} &1.013& -0.497& \textbf{0.000}\\
    %  \midrule
    %     \multicolumn{7}{c}{ATCE ($\lambda_1=1.5$) - (Test Acc:ERM: 80.75\%, CREDO: 79.30\%)}\\
    % \midrule
    % Latitude&0.335& \textbf{0.021}& 0.461& \textbf{0.505}& 0.090 &\textbf{0.999}\\
    %  O3&0.592 &\textbf{0.023}& 0.063& \textbf{1.004}& -0.740 &\textbf{1.000}\\
    %  SO2&0.823& \textbf{0.033}& \textbf{3.110} &1.456& 0.984& \textbf{1.000}\\
    % % Avg. &0.583& \textbf{0.026}& \textbf{1.211} &0.989 &-0.740 &\textbf{0.999}\\
     \bottomrule
    \end{tabular}
    }
    \caption{Results on SANGIOVESE}
    \label{tab:sangiovese_mehra}
\end{table}

% Our results on MEHRA, SACHS and Asia datasets from the BNLearn repository showed similar trends, and are reported in the Appendix. 
\vspace{-7pt}
\subsection{Synthetic Data With Non-Linear Domain Priors}
\label{synthetic_non_linear}
\begin{figure}
    \centering
    \includegraphics[width=.42\textwidth]{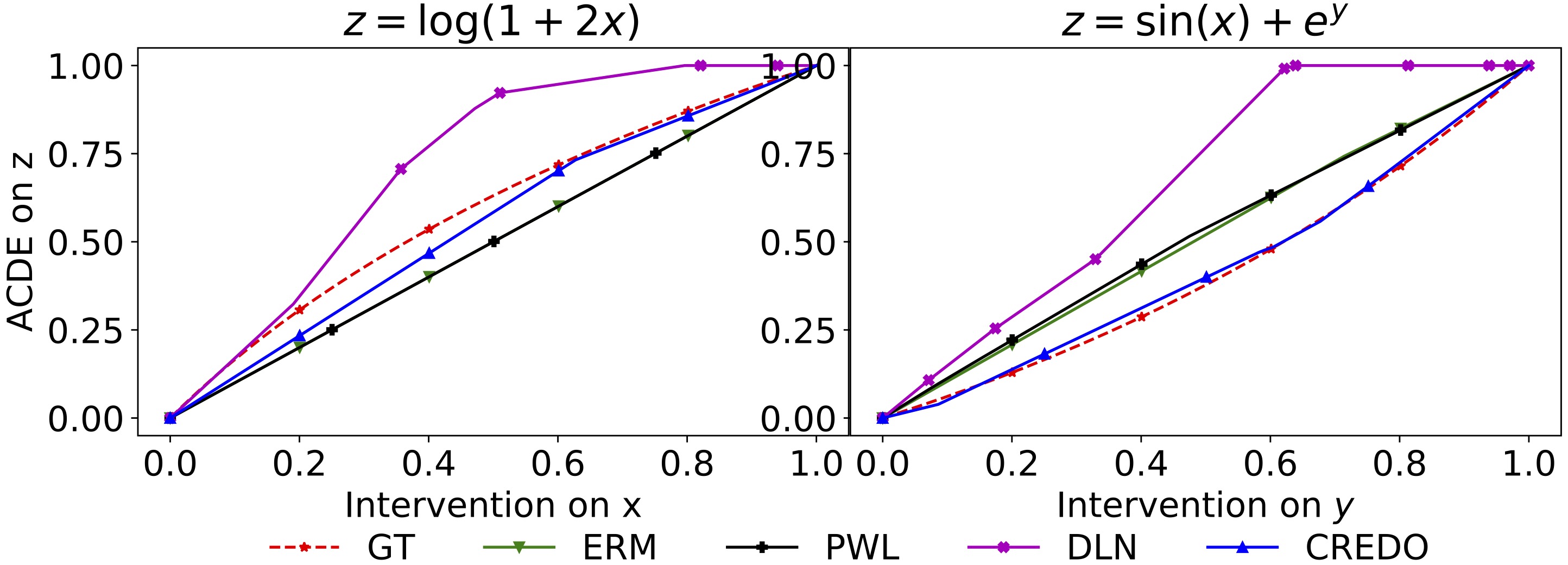}
    \caption{Enforcing monotonicity. ACDE plots of ERM, CREDO, PWL and DLN on Synthetic Tabular datasets. CREDO matches GT better than other methods.}
    \label{fig:synthetic_tabular}
\end{figure}

\vspace{-6pt}
For a fair comparison to PWL and DLN, we consider two synthetic datasets motivated from their paper. These datasets have non-linear monotonic relationships in the true structural equations, and we hence input a non-linear prior to CREDO. 
These datasets have the form $z=\log(1+2x), x\in[0,1]$ and $z=\sin(x)+e^y$, $x\in[0,1]$, $y\in[0,1]$. The domain priors provided by a user is a log-linear monotonic relationship of $x$ to output $z$ for $z=\log(1+2x), x\in[0,1]$ and $z$ is monotonically increasing w.r.t. $y$ when $x$ is kept constant for $z=\sin(x)+e^y$, $x\in[0,1]$, $y\in[0,1]$. CREDO uses the shape of the prior while PWL and DLN can only regularize for monotonic constraints. As a result, we observe that CREDO matches the prior shape better when compared to PWL and DLN (Fig \ref{fig:synthetic_tabular}). In particular, using DLN is worse than using ERM as the DLN model exaggerates the ACDE substantially (has higher RMSE and Frechet Score). In addition, there is no benefit in using PWL over ERM; both are almost identical in their ACDE. CREDO model is the closest to the ground-truth prior.
\vspace{-6pt}
\subsection{When Causal Graph is Unknown}
\label{subsec_expts_setting1}
As stated in Sec \ref{sec_identify_regularize}, when we do not know the causal graph, the best we can do is regularize for ACDE. Below, we apply CREDO on real-world datasets for regularizing ACDE of an NN. In such cases, we expect a domain expert to provide prior function properties such as shape and/or a search space for its parameters. %  we make the assumption that the parametric form of the prior function as well as the search spaces are given (which is reasonable to assume in this context). 
As stated earlier (and described in the Appendix), we find the best parameters for the prior function by tuning for highest validation-set accuracy. %More details about this procedure are in the Appendix.
\begin{figure}%[16]{r}{0.6\linewidth}
    \centering
    \includegraphics[width=0.47\textwidth]{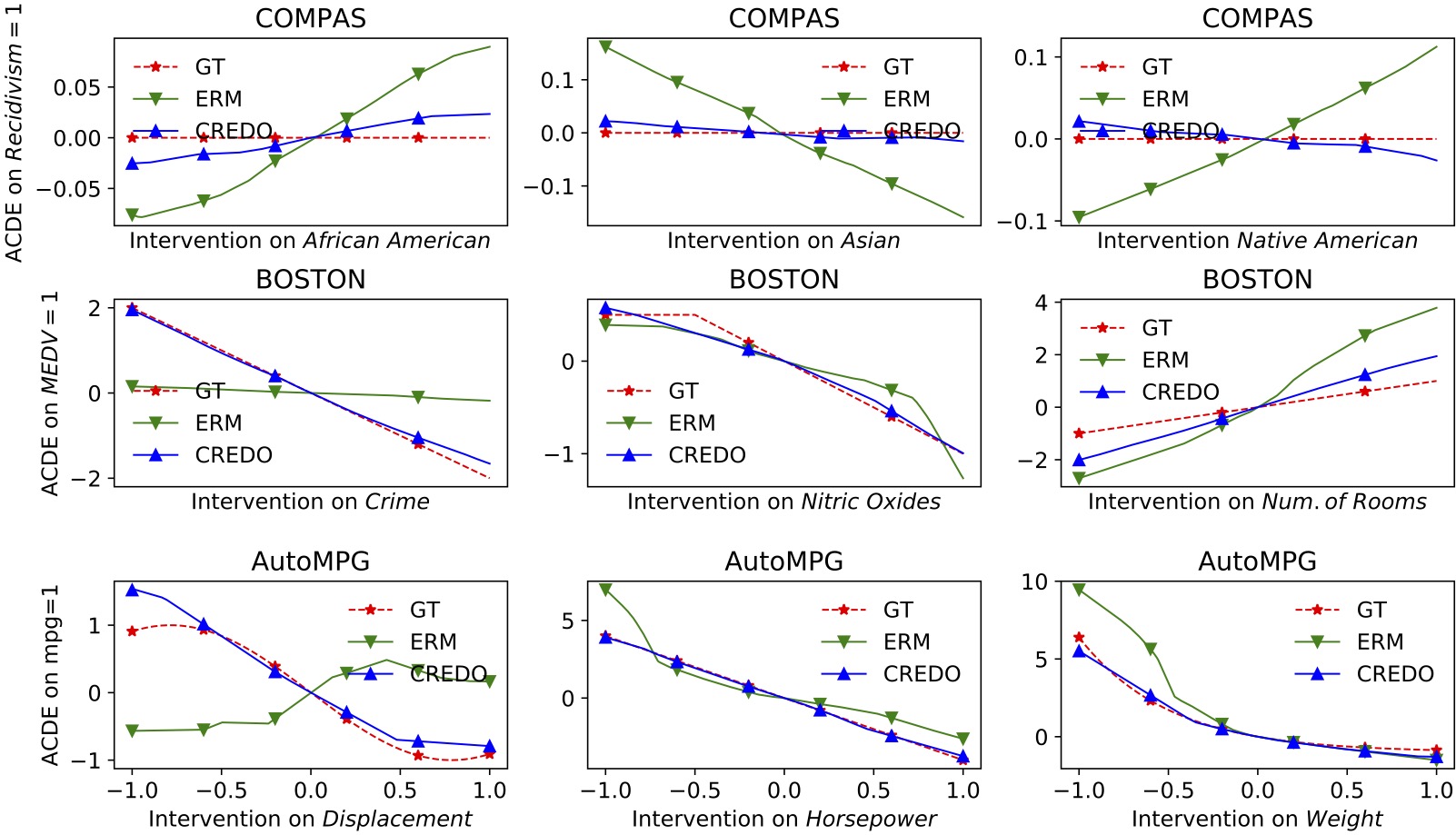}
    \caption{ Results on COMPAS, BOSTON, and AutoMPG datasets}
    \label{fig:auto_compas_boston}
    \vspace{-9pt}
\end{figure}

\vspace{-3pt} \noindent \textbf{COMPAS:}
%In this dataset%(Correctional Offender Management Profiling for Alternative Sanctions) 
The task herein~\cite{compas} is to predict the likelihood of \textit{two-year recidivism} (re-offending in next two years) given a set of features %This dataset is studied in fairness literature
\cite{fairness_crimilal_justice,berk2019accuracy,brennan2013emergence,ferguson2014big}. From a fairness perspective, we expect NN models to have zero direct causal effect of \textit{race} while predicting \textit{recidivism}~\cite{pearl2009causality}. Table \ref{tab:auto_compas_boston} and Fig \ref{fig:auto_compas_boston} shows our results. CREDO is able to align ACDE of \textit{race} on \textit{two-year recidivism} prediction to be close to zero with almost the same model accuracy. Note the significant reduction in RMSE and Frechet scores, which measure the alignment of the trained NN's ACE w.r.t. the domain prior. 

\vspace{-3pt} \noindent \textbf{AutoMPG:}
The AutoMPG dataset contains the \textit{mileage} of various cars given attributes such as \textit{weight}, \textit{horsepower}, \textit{displacement}, etc~\cite{Dua:2019}. We learn an NN classifier that can predict if the \textit{mileage} is better than average. We want \textit{displacement}, \textit{weight} and \textit{horsepower} to have a decreasing causal effect w.r.t. \textit{mileage}. 
Results shown in %Fig \ref{fig:auto_compas_boston} and 
Table \ref{tab:auto_compas_boston} and Fig \ref{fig:auto_compas_boston} once again support the usefulness of our method.

\vspace{-3pt} \noindent \textbf{Boston Housing:}
This dataset concerns home values in the suburbs of Boston~\cite{Dua:2019}. We convert the output attribute \textit{home value} into a categorical output, and learn a classifier that can predict if the home value is better than average. As the domain prior, we want \textit{crime rate} and \textit{concentration of Nitric Oxides} (above a threshold) to have a decreasing causal effect and \textit{number of rooms} (RM) to have an increasing causal effect w.r.t. housing prices. %Fig \ref{fig:auto_compas_boston} and 
Results in Table \ref{tab:auto_compas_boston} and Fig \ref{fig:auto_compas_boston} show that CREDO helps align the ACDE of the trained model with the prior in all these cases.
\begin{table}
\scalebox{0.65}{
\begin{tabular}{lcccccc}
\toprule
 Feature & \multicolumn{2}{c}{RMSE} & \multicolumn{2}{c}{Frechet Score} & \multicolumn{2}{c}{Corr. Coeff.}\\
 \midrule
 &ERM&CREDO&ERM&CREDO&ERM&CREDO\\
 \midrule
 \multicolumn{7}{c}{COMPAS ($\lambda_1=5$) (ERM test accuracy is 67.90\%, CREDO test accuracy is 67.09\%)}\\
 \midrule
 African American&0.055&\textbf{0.016}&0.088&\textbf{0.025}&-&-\\
 Asian&0.092&\textbf{0.018}&0.162&\textbf{0.021}&-&-\\
 Native American&0.059&\textbf{0.011}&0.109&\textbf{0.025}&-&-\\
 %Avg.&0.069&\textbf{0.012}&0.120&\textbf{0.024}&-&-\\
  \midrule
%   \multicolumn{7}{c}{Boston ($\lambda_1=2.2$) (ERM test accuracy is 88.2\%, CREDO test accuracy is 85.30\%)}\\
%  \midrule
%  Crime&0.52& \textbf{0.145}& \textbf{0.181}& 1.951& 0.996& \textbf{0.999}\\
%  Nitric Oxide&0.165& \textbf{0.080}& 1.265& \textbf{0.994}& 0.957& \textbf{0.991}\\
%  Num. of Rooms&0.994& \textbf{0.036}& 3.786& \textbf{2.009}& 0.993& \textbf{1.000}\\
%  %Avg.(Min. for Corr. Coeff.)&0.740& \textbf{0.080}& 1.744& \textbf{1.651}& 0.957& \textbf{0.991}\\
%  \midrule
 \multicolumn{7}{c}{AutoMPG ($\lambda_1=1.5$) (ERM test accuracy is 88.6\%, CREDO test accuracy is 87.34\%)}\\
 \midrule
 Displacement&1.144&\textbf{0.212}&\textbf{0.566}&1.524&-0.945&\textbf{0.977}\\
 Horsepower&1.036&\textbf{0.081}&6.978&\textbf{3.908}&0.922&\textbf{0.999}\\
 Weight&1.780&\textbf{0.25}&9.453&\textbf{5.510}&0.986&\textbf{0.992}\\
 %Avg.(Min. for Corr. Coeff.)&1.181&\textbf{0.109}&4.20&\textbf{2.189}&-0.945&\textbf{0.977}\\
  \midrule
  \multicolumn{7}{c}{Boston ($\lambda_1=2.2$) (ERM test accuracy is 88.2\%, CREDO test accuracy is 85.30\%)}\\
 \midrule
 Crime&0.52& \textbf{0.145}& \textbf{0.181}& 1.951& 0.996& \textbf{0.999}\\
 Nitric Oxide&0.165& \textbf{0.080}& 1.265& \textbf{0.994}& 0.957& \textbf{0.991}\\
 Num. of Rooms&0.994& \textbf{0.036}& 3.786& \textbf{2.009}& 0.993& \textbf{1.000}\\
\bottomrule
\end{tabular}
}
\caption{Enforcing Causal Effects (ACDE) of Multiple Variables: Results on COMPAS, AutoMPG, and BOSTON datasets}
\vspace{-9pt}
\label{tab:auto_compas_boston}
\end{table}
We obtain similar results on enforcing domain priors over the MEPS, Law School Admission, Adult, Car Evaluation, and Titanic datasets, as shown in the Appendix. 
\vspace{-7pt}
\subsection{Robustness to Noisy Input}
\vspace{-6pt}
%\noindent \textbf{Robustness to Noisy Input.}
We study the robustness of CREDO models when the test data is noisy. For SANGIOVESE and AutoMPG (classification datasets), we perturb test samples by adding zero-mean Gaussian noise. For synthetic tabular (regression) datasets, we study robustness to out-of-domain test samples (the domains are given in section~\ref{synthetic_non_linear}). As in Table~\ref{tab:input_robustness}, CREDO models, which are trained to respect true causal relationships, are more robust to test-time perturbations (especially with higher noise variance) for both classification and regression datasets. 
\begin{table}
    \centering
    \scalebox{0.58}{
    \begin{tabular}{lllllll}
        \toprule
        &\multicolumn{2}{c}{Variance 0.2}&\multicolumn{2}{c}{Variance 0.5}&\multicolumn{2}{c}{Variance 1.0}\\
        \midrule
        Dataset &ERM&CREDO&ERM&CREDO&ERM&CREDO\\
        \midrule
        SANGIOVESE&73.15&\textbf{73.95}&61.60&\textbf{63.20}&55.45&\textbf{56.00}\\
        AutoMPG&\textbf{86.07}&84.80&79.74&\textbf{83.54}&68.35&\textbf{79.74}\\
        \midrule
         Dataset &\multicolumn{3}{c}{ERM}&\multicolumn{3}{c}{CREDO}\\
        \midrule
        $z=\log(1+2x), x\in[1,2]$&\multicolumn{3}{c}{0.094}&\multicolumn{3}{c}{\textbf{0.073}}\\
        $z=\sin(x)+e^y$, $x\in[0,1]$, $y\in[1,2]$&\multicolumn{3}{c}{21.75}&\multicolumn{3}{c}{\textbf{20.29}}\\
        \bottomrule
    \end{tabular}
    }
    \caption{Robustness results: Accuracy on noisy test data and MSE on out-of-distribution samples of ERM, CREDO models}
    \label{tab:input_robustness}
    \vspace{-7pt}
\end{table}
\vspace{-7pt}
\subsection{Relation to Fairness}
\vspace{-6pt}
Regularizing ACDE can be viewed as being close to achieving \textit{no proxy discrimination in expectation}: $\mathbb{E}(\hat{Y}|do(T=t)) = \mathbb{E}(\hat{Y}|do(T=t'))$~\cite{disc_neurips} in fairness applications. That is, NN output $\hat{Y}$ should not depend on the intervention on $T$ when making predictions. To encourage fairness, we enforce the ACDE of protected attributes on the outcome to be zero so that all interventions to the treatment variable $T$ lead to the same outcome, zero in this case. Results shown in the Table~\ref{fairness} demonstrate that CREDO outperforms ERM on various datasets w.r.t. the fairness metric: \textit{Disparate Impact} (higher is better) which captures the property of \textit{no proxy discrimination in expectation} through the ratio $\frac{p(\hat{Y}=1|do(T=t))}{p(\hat{Y}=1|do(T=t'))}$. Where $\hat{Y}=1$ signifies positive outcome, $t$ denotes the unprivileged group, and $t'$ denotes the privileged group.
\begin{table}%[5]{r}{0.65\linewidth}
\footnotesize
    \centering
    \scalebox{0.65}{
    \begin{tabular}{ccccccc}
    \toprule
    Datasets ($\rightarrow$)&\multicolumn{2}{c}{Law School}& MEPS&\multicolumn{3}{c}{COMPAS}\\
    \midrule
        Model ($\downarrow$) & Gender & Race & Race &African American&Asian&Native American\\
         \midrule
         ERM & 0.94 & 0.91&0.77&0.82&\textbf{0.95}&0.63\\
         CREDO & \textbf{0.99} & \textbf{0.94}&\textbf{0.84}&\textbf{0.83}&0.84&\textbf{0.82}\\
         \bottomrule
    \end{tabular}
    }
    \caption{ERM vs CREDO on \textit{Disparate Impact} metric}
    \label{fairness}
    \vspace{-9pt}
\end{table}
We also compared CREDO with two standard fair classification algorithms: \textit{Exponentiated Gradient (EG)} and \textit{Grid Search (GS)}~\cite{reductions_approach}(we use the code from \url{https://fairlearn.org} to implement these algorithms) on the Boston Housing dataset (Table~\ref{tab_reductions}). Along with ACDE, we use a fairness-based metric: \textit{Demographic Parity Difference} (Lower is better). We enforced a constraint that ACDE of sensitive features: \textit{proportion of black people} (B) and \textit{\% of lower status of population} (LSTAT) should be zero on house price prediction.
\begin{table}[H]%[8]{r}{0.67\linewidth}
\footnotesize
    \centering
    \scalebox{0.9}{
    \begin{tabular}{ccccc}
    \toprule
    Model&ACDE&ACDE&Demog&Test\\
    &(B)&(LSTAT)&Par Diff&Accuracy\\
    \midrule
        ERM &0.10&0.39&0.86&88.23\%\\
        EG & 0.18&0.33 & \textbf{0.45}&77.50\%\\
        GS & 0.08 & 0.33 & 0.79 & 86.50\%\\
        CREDO&\textbf{0.00} & \textbf{0.23} & 0.66 & 86.27\%\\
    \bottomrule
    \end{tabular}
    }
    \caption{CREDO vs fair classifiers}
    \label{tab_reductions}
    \vspace{-9pt}
\end{table}
Results show that CREDO performs on par w.r.t. fairness metric and better w.r.t. ACDE metric, which captures causal attribution. 

Results on more datasets are in Appendix~\ref{app:results}. Ablation studies, effect of regularization co-efficient, time complexity,   and discussion on incorrect priors are in Appendix~\ref{sec_ablations}.
\vspace{-7pt}
\section{Conclusion}
\vspace{-6pt}
\label{sec_conclusions}
In this work, we propose a new causal regularization method, CREDO, that can learn neural network models whose learned causal effects match prior domain knowledge as provided by an expert user. We show that the method can work with any differentiable prior representing complete or partial understanding of the domain. Importantly, we distinguish between direct and total causal effects, and show how both can be considered in CREDO. %We formally analyze our method and show why the regularizer helps align the learned causal effect with the prior. 
We performed extensive experiments on various datasets, with known and unknown causal graphs, and with different kinds of priors.  CREDO shows promising performance in matching causal domain priors while maintaining test accuracy.
\vspace{-7pt}
\section*{Acknowledgements}
\vspace{-6pt}
We are grateful to the Ministry of Education, India for the financial support of this work through the Prime Minister's Research Fellowship (PMRF) and UAY programs. We thank the anonymous reviewers for their valuable feedback that helped improve the presentation of this work.
\clearpage
\bibliography{bibliography}
\bibliographystyle{icml2022}

\clearpage
\appendix
\section*{Appendix}
In this appendix, we include the following information.
\begin{itemize}
    \item Proofs of propositions in Section~\ref{proofs}
    \item Discussion on sources of causal domain priors for using CREDO in practice in Section~\ref{app:priors}
    \item Implementation details of CREDO in Section~\ref{implementation}
    \item More experimental results in Section~\ref{app:results} 
    \begin{itemize}
        \item Experiments on BNLearn datasets
        \item Enforcing fairness constraints
    \end{itemize}
    \item Ablation studies and analysis in Section ~\ref{sec_ablations}
    \item Description of computation of ACE (Average Causal Effect), as in \cite{chattopadhyay2019neural} in Section~\ref{acealgo}
    \item Causal graphs/DAGs for BNLearn datasets in Section~\ref{dags}
    \item Architectures/training details of our models for all datasets in Section~\ref{arch}
\end{itemize}
\vspace{-7pt}
\section{Proofs of Propositions}
\label{proofs}
\vspace{-6pt}
We begin writing the proofs of our propositions by recollecting two key results from~\cite{pearl2009causality} which we use in our proofs: (i) When there is no backdoor path from treatment $T$ to the outcome $\hat{Y}$, interventional distribution is equal to the conditional distribution i.e., $p(\hat{Y}|do(T=t)) = p(\hat{Y}_t) = p(\hat{Y}|T=t)$. (ii) If there exist a set of nodes $W$ that satisfy the backdoor criteria relative to the causal effect of $T$ on $\hat{Y}$, the causal effect of $T$ on $\hat{Y}$ can be evaluated using the adjustment formula $\mathbb{E}[\hat{Y}_t] = \mathbb{E}[\hat{Y}|do(T=t)] = \sum_{\hat{Y}}\hat{Y}p(\hat{Y}|do(T=t)) = \sum_w \sum_{\hat{Y}} \hat{Y}p(\hat{Y}|T=t,W=w)p(W=w)$. Note that this is equivalent to $\mathbb{E}_{W}[\hat{Y}|t,W]$ in our notation, since the inner expectation over $\hat{Y}$ vanishes due to the deterministic nature of the NN.  %, i.e. $\sum_w p(\nabla_t \hat{Y}|T=t,W=w)p(W=w) =\mathbb{E}_{W}[\nabla_t \hat{Y}|t,W] $.)%Note that, in our proofs, we use the notation $\frac{\partial^n [\hat{Y}|t,z,w]}{\partial t^n}$ to refer to the quantity $\frac{\partial^n [\hat{Y}]}{\partial t^n}$ evalulated at the point $(t,z,w)$.
\acdeidentifiability*
\vspace{-3pt}
\begin{proof}
Starting with the definition of $ACDE$ of $T$ at $t$ on $\hat{Y}$ (Equation 1 of main paper), we get
\begin{equation*}
\label{cde_eq1}
\begin{aligned}
    ACDE^{\hat{Y}}_{t} &= \mathbb{E}_{{Z,W,U}}[\hat{Y}_{t,{Z,W}}]-\mathbb{E}_{{Z,W,U}}[\hat{Y}_{t^*,{Z,W}}]]\\
    &=\mathbb{E}_{Z,W}[\hat{Y}_{t,{Z,W}}]-\mathbb{E}_{Z,W}[\hat{Y}_{t^*,{Z,W}}\big]\\
    &=\mathbb{E}_{Z,W}[\hat{Y}|t,{Z,W}]-\mathbb{E}_{Z,W}[\hat{Y}|t^*,{Z,W}\big]
\end{aligned}
\end{equation*}
Since a NN $f$ is a deterministic function of its inputs, expectation over $U$ can be discarded once we condition on all features (second equality above). Further, once all parents of $\hat{Y}$ (i.e., $T,Z,W$) are intervened, there cannot be an unobserved confounder that causes both $\hat{Y}$ and $T$. Hence unconfoundedness is valid for the effect from $T$ to $\hat{Y}$. For this reason, we can replace the intervention with conditioning in the last step.
\end{proof}
\acderegularization*
\vspace{-3pt}
\begin{proof}
Using the identifiability result from Proposition \ref{proposition:proposition_cde_identifiability},
$$ACDE^{\hat{Y}}_{t} = \mathbb{E}_{{Z,W}}\big[\hat{Y}|t,{Z,W}\big]-\mathbb{E}_{{Z,W}}\big[\hat{Y}|t^*,{Z,W}]\big]$$
Now, taking the $n^{th}$ partial derivative w.r.t $t$ on both sides,
\begin{align*}
\frac{\partial^n ACDE^{\hat{Y}}_{t}}{\partial t^n} &= \frac{\partial^n [\mathbb{E}_{{Z,W}}[\hat{Y}|t,{Z,W}]-\mathbb{E}_{{Z,W}}[\hat{Y}|t^*,{Z,W}]]}{\partial t^n}\\
     &= \frac{\partial^n [\mathbb{E}_{{Z,W}}[\hat{Y}|t,{Z,W}]]}{\partial t^n} (\because t^*\ is\ a\ constant)\\
     &= \mathbb{E}_{{Z,W}} \left[\frac{\partial^n [\hat{Y}({t,{Z,W}})]}{\partial t^n}\right]
\end{align*}
Note the change in notation in last step ($\hat{Y}({t,{Z,W}})$ vs $\hat{Y}|{t,{Z,W}}$). This change is merely to differentiate between conditioning (when taking expectation) and evaluating (when taking partial derivative). However both quantities are the same.
\end{proof}
\andeidentifiability*
\vspace{-3pt}
\begin{proof}
Assuming that the set $W$ forms a valid adjustment set in the calculation of causal effect of $T$ on $\hat{Y}$, from the backdoor adjustment formula~\cite{pearl2009causality}, 
we get $\mathbb{E}_{U}[\hat{Y}_{t,Z_{t^*}}]=\mathbb{E}_{W,U} [\hat{Y}_{t,Z_{t^*}}|W]$. However, the value $Z_{t^*}$ not only depends on the intervention $do(T=t^*)$ but also on $U, W$. That is, $Z_{t^*}$ is a random variable which is a function of $W,U$. So, $\mathbb{E}_{W,U} [\hat{Y}_{t,Z_{t^*}}|W]$ can be written as~\cite{pearl2001direct}:
\begin{equation*}
    \mathbb{E}_{W,U} [\hat{Y}_{t,Z_{t^*}}|W] = \mathbb{E}_{W,U}\sum_{Z_{t^*}} [\hat{Y}_{t, Z_{t^*}}|Z_{t^*},W]p(Z_{t^*}|W)
\end{equation*}
Using \textit{consistency}~\cite{pearl2009causality} and \textit{backdoor} adjustment, we have:
$\mathbb{E}_{U}[\hat{Y}_{t,Z_{t^*}}]=\mathbb{E}_{Z_{t^*},W,U} [\hat{Y}|{t,Z_{t^*},W}]$.
%we get $\mathbb{E}_{U}[\hat{Y}_{t,Z_{t^*}}]=\mathbb{E}_{Z_{t^*},W,U} [\hat{Y}|{t,Z_{t^*},W}]$.We get the expectation over $Z_{t^*}$ because the value $Z_{t^*}$ not only depends on the intervention $do(T=t)$ but also on $U, W$~\cite{pearl2001direct}.
\begin{equation*}
    \label{eq:nde}
    \begin{aligned}
        ANDE^{\hat{Y}}_{t} &= \mathbb{E}_{U}[\hat{Y}_{t,Z_{t^*}}]-\mathbb{E}_{U}[\hat{Y}_{t^*,Z_{t^*}}]\\
        &=\mathbb{E}_{Z_{t^*},W,U}[\hat{Y}|t,Z_{t^*},W]-\mathbb{E}_{Z_{t^*},W,U}[\hat{Y}|t^*,Z_{t^*},W]]\\
        &=\mathbb{E}_{Z_{t^*},W}[\hat{Y}|t,Z_{t^*},W]-\mathbb{E}_{Z_{t^*},W}[\hat{Y}|t^*,Z_{t^*},W]
    \end{aligned}
\end{equation*}
The second equality is because of the adjustment formula as described above. Further, similar to the ACDE case, the third equality is because once we condition on all input features, $f$ is a deterministic function of its inputs and hence expectation over noise variables can be omitted. Further, $Z_{t^*}$ is identified because all parents of Z are observed as per Assumption 1. 
\end{proof}
\anderegularization*
\begin{proof}
Using the identifiability result from Proposition \ref{proposition:proposition_nde_identifiability},
$$ANDE^{\hat{Y}}_{t} = \mathbb{E}_{Z_{t^*},W}[\hat{Y}|t,Z_{t^*},W] - \mathbb{E}_{Z_{t^*},W}[\hat{Y}|t^*,Z_{t^*},W]$$
Now, taking the $n^{th}$ partial derivative w.r.t $t$ on both sides,
\begin{align*}
    \frac{\partial^n ANDE^{\hat{Y}}_{t}}{\partial t^n} &= \frac{\partial^n \big[\mathbb{E}_{Z_{t^*},W}[\hat{Y}|t,Z_{t^*},W] - \mathbb{E}_{Z_{t^*},W}[\hat{Y}|t^*,Z_{t^*},W]\big]}{\partial t^n}\\
     &= \frac{\partial^n \big[\mathbb{E}_{Z_{t^*},W}[\hat{Y}|t,Z_{t^*},W]\big]}{\partial t^n}(\because t^*\ is\ a\ constant)\\
     &= \mathbb{E}_{Z_{t^*},W} \left[\frac{\partial^n [\hat{Y}(t,Z_{t^*},W)]}{\partial t^n}\right]
\end{align*}
\end{proof}
\atceidentifiability*
\begin{proof}
Assuming that the set $W$ forms a valid adjustment set in the calculation of causal effect of $T$ on $\hat{Y}$, similar to the proofs above, from the backdoor adjustment formula~\cite{pearl2009causality}, we get:
\begin{equation*}
    \label{eq:tce}
    \begin{aligned}
        ATCE^{\hat{Y}}_{t} &= \mathbb{E}_{U}[\hat{Y}_{t,Z_{t}}-\hat{Y}_{t^*,Z_{t^*}}]\\
        &=\mathbb{E}_{Z_{t},W,U}[\hat{Y}|t,Z_{t},W]-\mathbb{E}_{Z_{t^*},W,U}[\hat{Y}|t^*,Z_{t^*},W]]\\
        &=\mathbb{E}_{Z_{t},W}[\hat{Y}|t,Z_{t},W]-\mathbb{E}_{Z_{t^*},W}[\hat{Y}|t^*,Z_{t^*},W]
    \end{aligned}
\end{equation*}
Similar to ANDE identifiability in Proposition~\ref{proposition:proposition_nde_identifiability}, we can reason about the explicit expectation over $Z_t, Z_{t^*}$ in the second equality. Similar to the ACDE, ANDE identifiability in Propositions~\ref{proposition:proposition_cde_identifiability},~\ref{proposition:proposition_nde_identifiability}, once we condition on all input features, $f$ is a deterministic function of its inputs and hence expectation over noise variables can be omitted.
\end{proof}
Note that the values of $Z_t, Z_{t^*}$ in both cases (ANDE, ATCE) are decided by the causal mechanism that generates $Z$ which are learned as separate regressors in our implementation.
\atceregularization*
\begin{proof}
Let $f^{i}$ denote the structural equation of the underlying causal model for variable $Z^i$ (a total of $K$ such equations, which are learned separately). W.l.o.g., assuming $Z^i$s are topologically ordered, we have $Z^i=f^{i}(T,Z^1,Z^2,\dots,Z^{i-1},W)$ (we add W for generality; it is not necessary for all variables in $W$ to cause $Z^i$). Consequently, at $T=t$, we have $Z^i_{t}=f^{i}(t,Z^1_{t},Z^2_{t},\dots,Z^{i-1}_{t},W)$. Consider the first-order Taylor expansion of $Z^k_{t+\Delta t}$:
{\small
\begin{equation}
\label{eq:taylors}
    \begin{aligned}
    Z^k_{t+\Delta t} \approx Z^k_{t} + \Delta t \bigg[\frac{df^k}{dt}\bigg],
    \text{ where }\ \frac{df^k}{dt} = \frac{\partial f^k}{\partial t} + \sum_{j=1}^{k-1} \frac{\partial f^k}{\partial Z^j}\frac{df^j}{dt} 
    \end{aligned}
\end{equation}
}
\noindent We can prove Eqn \ref{eq:taylors} by induction over $k$. Since we are interested in the behavior of TCE w.r.t. the interventional value, we consider the first-order Taylor expansion of $\hat{Y}_{t+\Delta t}$
{\small
\begin{equation}
\label{eq:tce_taylor}
\begin{aligned}
    \hat{Y}_{t+\Delta t} &= f(t+\Delta t, Z^{1}_{t+\Delta t}, \dots, Z^{K}_{t+\Delta t}, W)\\
    &\approx f(t+\Delta t, Z^1_{t}+\Delta t[\frac{df^1}{dt}], \dots, Z^K_{t}+\Delta t\bigg[\frac{df^K}{dt}\bigg], W)\\
    &\approx \hat{Y}_{t} + \Delta t\bigg[\frac{\partial f}{\partial t}+ \sum_{j=1}^{K}\frac{\partial f}{\partial Z^j}\frac{df^j}{dt} \bigg]\\
\end{aligned}
\end{equation}
}
\noindent Taking $\hat{Y}_{t}$ to the left, and adding-subtracting $\hat{Y}_{t^*}$, we get the LHS of Equation \ref{eq:tce_taylor} as $\Delta TCE^{\hat{Y}}_{t}=TCE^{\hat{Y}}_{t+\Delta t}-TCE^{\hat{Y}}_{t}$. In the limit that the perturbation $\Delta t$ is very small, we get rid of the error introduced by the first-order Taylor approximations. Thus we have:
{\small
\begin{equation}
\label{equation9}
    \lim_{\Delta t\to 0} \frac{\Delta TCE^{\hat{Y}}_{t}}{\Delta t} \equiv \frac{d TCE^{\hat{Y}}_{t}}{dt} = \frac{\partial f}{\partial t}+ \sum_{j=1}^{K}\frac{\partial f}{\partial Z^j}\frac{df^j}{dt}
\end{equation}
}
\noindent Finally, taking expectation on both sides completes the proof (by Leibniz integral rule).
$$\frac{d ATCE^{\hat{Y}}_{t}}{dt} = \mathbb{E}_{Z_t,W} \left[\frac{d [\hat{Y}(t,Z_{t},W)]}{dt}\right] $$
\begin{align*}
    \text{ where }\ \frac{d [\hat{Y}(t,Z_{t},W)]}{dt}&=\frac{\partial [\hat{Y}(t,Z_{t},W)]}{\partial t}\\
    &+ \sum_{j=1}^{K}\frac{\partial [\hat{Y}(t,Z_{t},W)]}{\partial Z^j}\frac{d[f^j|t,Z_{t},W]}{dt}
\end{align*}
\end{proof}

\section{Availability of Causal Domain Priors}
\label{app:priors}
Our work is based on priors for the causal effect of a feature provided by domain experts. Extending our discussion on the availability of such causal domain priors in the main paper, we provide below a more detailed discussion and list examples of different kinds of priors that are commonly known in fields such as algorithmic fairness, economics, health  and physical sciences, and can be used in CREDO for building robust prediction models. A summary of different kinds of domain priors is presented in Table~\ref{tab:domain_priors} of main paper.
\begin{itemize}
    \item \underline{\textit{U-shaped causal effects:}} There are many situations where it is known that a feature's effect follows a \textit{U-shaped} pattern, i.e. increasing the feature may increase the outcome up to a point, after which it starts decreasing the outcome. In medicine, U-shaped curves have been found for various factors, such as cholesterol level, diastolic blood pressure~\cite{salkind-ushaped2010}, and the dose-response curve for drugs~\cite{calabrese2001u}. Another example is the relationship of mortality with age for certain diseases, where infants and elderly may experience the highest mortality (although sometimes more complex relationships are observed)~\cite{luk2001mortality}. All such effects can be modeled by the proposed CREDO algorithm.  
    \item \underline{\textit{J-shaped causal effects:}} \cite{fraser_nonlinear_2016} studied the J-shaped relationship between exposure and outcome, such as between alcohol with coronary heart disease. Similarly, \cite{flanders_adjusting_2008} observed a J-shaped relationship between mortality and Body Mass Index. CREDO allows the use of such non-linear priors while training a model and thus retain these causal relationships in the NN model. %\item \underline{\textit{General functions:}} As another example, consider the charging voltage curve for a battery that shows how the voltage increases with charging time~\cite{han2019compare-battery}. Accurate prediction of battery voltage is required in many scenarios, e.g., activity planning for drones. These curves follow a characteristic pattern based on battery type (e.g., higher rate at beginning and end, and lower rate in the middle), and each such curve can be provided as a prior for robust prediction using the CREDO algorithm. 
    \item \underline{\textit{Zero (direct) causal effect:}} In many cases, a feature may have a spurious correlation  with the outcome and an ML model should not learn such correlations. For example, skin color should not matter in a classifier based on facial images~\cite{dash2022imagecf}, race/caste should not matter in a loan approval prediction model in a bank, and so on. A special case comes up in algorithmic fairness where we may allow total effect of a sensitive feature to be non-zero, but require the direct effect to be zero~\cite{Zhang2018FairnessID}. For example,  sensitive demographic features may affect a college admission decision mediated by the test score, but not directly. By formally distinguishing between direct and total effect regularization, our method makes it possible to regularize for such constraints.
    \item \underline{\textit{General monotonicity (e.g. ``diminishing returns''):}} In addition to plain monotonicity as done by \cite{gupta2019incorporate} and \cite{you2017deep}, domain experts often have additional information. The relationship may be super-linear (gradient of gradient is positive) or sub-linear. A special case of sub-linear monotonicity is the \textit{diminishing returns} observation~\cite{brue1993retrospectives} in economics (e.g., increasing the number of employees has a positive effect on productivity, but the effect decreases with the number of people already added). This is a popular submodular prior for many scenarios~\cite{kahneman2013choices} and can be enforced by our method, while prior methods on monotonicity do not provide any guarantees on specific functions. 
\end{itemize}

% \section{Assumption of Causal Inference}
% Our work relies on the assumptions of \textit{Unconfoundedness} and \textit{Positivity} (Section 3 of main paper). For simplicity and clarity, we used unconfoundedness assumption directly in the main paper. We now list a set of weaker assumptions/conditions than \textit{Unconfoundedness} which are enough to support all the theoretical analysis in this work.
% For the treatment $T$, whose effect we want to regularize, we assume that ``conditioning on all other features is sufficient to block the backdoor paths from $T$ to $\hat{Y}$''. This assumption holds true whenever one of the following conditions hold in the causal graph $\mathcal{G}$.
% \begin{itemize}
%     \item Whenever $T$ does not cause any other feature with whom it shares an unobserved confounding.
%     \item Whenever $T$ does not cause any other feature.
%     \item Whenever $T$ does not have any unobserved confounder w.r.t. $\hat{Y}$.
% \end{itemize}
\vspace{-7pt}
\section{Implementation Details of CREDO}
\label{implementation}
\vspace{-6pt}
\subsection{Evaluating $Z_t, Z_{t^*}$}
The definitions of ANDE and ATCE require the values $Z_t, Z_{t^*}$ to evaluate the causal effect of $T$ on $\hat{Y}$. To find $Z_t, Z_{t^*}$, we need to know the structure of the causal graph. If complete causal graph is not available, we at least need to know the partial causal graph involving $Z$. Since ANDE and ATCE are regularized for the setting where we have access to causal graph, we chose BNLearn datasets for our study: SANGIOVESE, MEHRA, ASIA (Lung Cancer), and SACHS. In these datasets, we model each structural equation of $Z^i\in Z$ as a function of its parents in the form of separate linear regressors. These regressors are learned independently from the model being trained and used in the equations for ANDE and ATCE.

\subsection{How is The Exact Functional Form of The Prior Determined?}
\label{app:priorsearch}
\begin{algorithm}
\DontPrintSemicolon
\footnotesize
\SetAlgoLined
\KwResult{Parameter sets of N prior functions: $\beta_1,\dots,\beta_n$}
\textbf{Input:} Domains $\mathcal{B}_1,\dots,\mathcal{B}_n$ of $\beta_1,\dots,\beta_n$, untrained NN $f$, $\mathcal{D}=\{(x^j, y^j)\}_{j=1}^N$, $y^j \in \{0,1,\dots,C\}$, $x^j \sim X^j$;.

\textbf{Initialize:} $i=I$, $best=0,\ accuracy=0, \beta_i\sim \mathcal{B}_i \ \forall i$

\While{$i > 0$}{
$(\beta_1',\dots,\beta_n') \sim (\mathcal{B}_1,\dots,\mathcal{B}_n)$

$accuracy$ =  $f_{\beta_1',\dots,\beta_n'}(\mathcal{D})$
\tcc{Algorithm~\ref{algo:regularizer}}

\If{$best<accuracy$}{
$best = accuracy$

$\beta_j = \beta_j'; \forall j\in[1,\dots,n]$
}

$i = i-1$
}
return $\beta_1,\dots,\beta_n$.
\caption{Prior function parameter search}
\label{algo:regularizer_2}
\end{algorithm}
If the exact functional form of the prior is provided, we can use it as it is in our method. However, we often get causal domain prior as a shape rather than an exact function. When the true parameters of such a function are not known, we search over possible values they can take (within a range) and choose the ones with the highest validation-set classification accuracy (this would function like any other hyperparameter search done for neural network models. see Algorithm~\ref{algo:regularizer_2}). For example, if the prior shape is linear w.r.t. a feature $t$, we search for a hyperparameter value for the slope $\alpha$ such that prior function is $\alpha t+c$ and choose the value with the highest validation accuracy. Performing a simple linear search over 10-20 values of $\alpha$ (e.g., 0.1 to 2 with 0.1 increment) suffices to achieve better/equal performance compared to ERM. We observed this in most of our experiments. For non-linear priors, a similar search can be performed as long as a parametric form and a reasonable search-space can be assumed. We, in general, assume that a prior shape (and/or a search space) are provided as the domain priors in this work.

\section{More Experimental Results}
\label{app:results}
\subsection{Enforcing Fairness Constraints}
\begin{figure}
\centering
\includegraphics[width=0.4\textwidth]{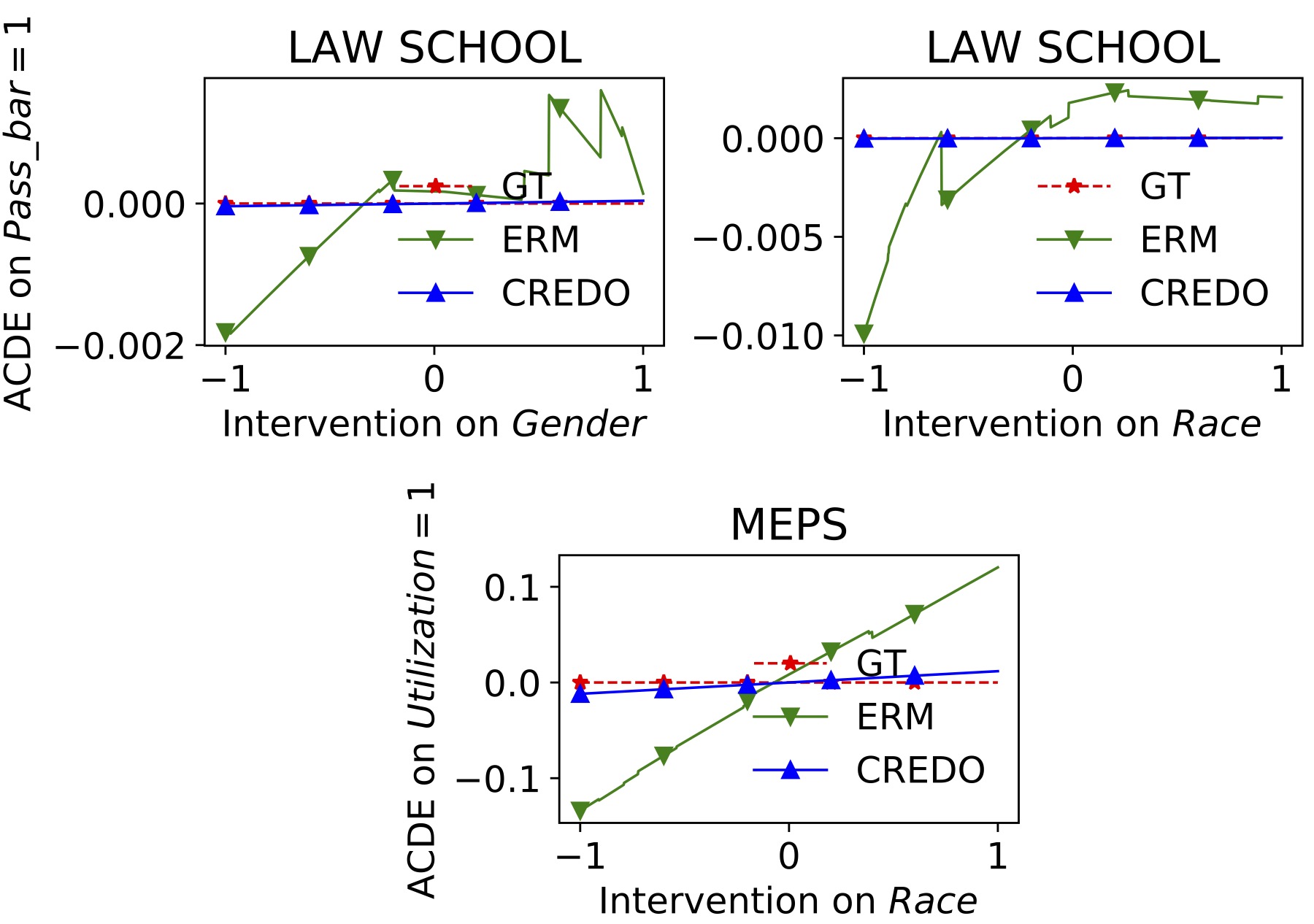}
\caption{ Enforcing Zero Causal Effect: Plot of ACDE of sensitive attributes on outcome in Law School, MEPS datasets.}
\label{fig:lawschool_meps}
\end{figure}
\paragraph{Law School Admission:}
In the Law School Admission dataset~\cite{lawschool}, the task is to predict whether a student gets admission into a law school based on a set of features. We expect a model to have no influence of protected attributes like \textit{gender,race} on \textit{admission}. Table \ref{tab:lawschool_meps} and Fig \ref{fig:lawschool_meps} reiterate our claims of CREDO's usefulness in forcing the ACDE of \textit{gender,race} on \textit{admission} to be zero without drop in model accuracy.
\begin{table}
    \centering
    \scalebox{0.9}{
    \begin{tabular}{lcccc}
        \toprule
         Feature&\multicolumn{2}{c}{RMSE}&\multicolumn{2}{c}{Frechet Score}\\
         \midrule
         &ERM&CREDO&ERM&CREDO\\
         \midrule
         \multicolumn{5}{c}{LAW-ERM test accr: 95.50\%, CREDO test accr: 95.39\%}\\
         \midrule
         Gender&$8e-4$&$\mathbf{2e-5}$&$1e-3$&$\mathbf{3e-5}$\\
         Race&$2e-3$&$\mathbf{1e-5}$&$1e-2$&$\mathbf{2e-5}$\\
         %Avg.&$1e-3$&$\mathbf{1.5e-5}$&$5e-3$&$\mathbf{2.5e-5}$\\
         \midrule
         \multicolumn{5}{c}{MEPS-ERM test accur: 86.27\%, CREDO test accur: 86.04\%}\\
         \midrule
         Race&0.02&\textbf{0.002}&0.03&\textbf{0.002}\\
         \midrule
    \end{tabular}
    }
    \caption{Results on Law School, MEPS datasets}
    \label{tab:lawschool_meps}
    \vspace{-7pt}
\end{table}
\vspace{-3pt} \noindent \textbf{MEPS:}
In the MEPS (Medical Expenditure Panel Survey) dataset~\cite{meps}, usually the task is to predict the \textit{utilization} score of an individual. Utilization can be thought of as requiring additional care for an individual, and such decisions should be made independent of the \textit{race} of the individual. Table \ref{tab:lawschool_meps} and Fig \ref{fig:lawschool_meps} once again show that CREDO is able to force the ACDE of \textit{race} on \textit{utilization} to be zero with insignificant effect on model accuracy.

\vspace{-3pt} \noindent \textbf{Adult:}
\begin{figure}
\centering
\includegraphics[width=0.45\textwidth]{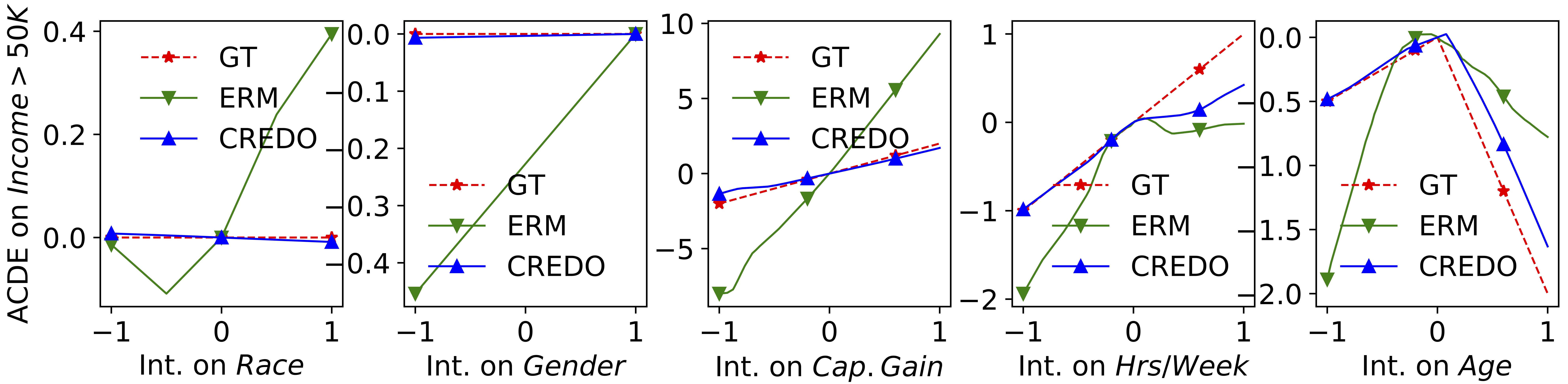}
\caption{\small Enforcing Monotonicity and Zero Causal Effect: Plot of causal effects learned by models trained on Adult dataset.}
\label{fig:adult}
\end{figure}
In this experiment, we study the effectiveness of CREDO under multiple constraints. The Adult dataset is a real-world dataset from the UCI repository~\cite{Dua:2019}. On the Adult dataset, we learn a classifier that can predict if the yearly income of an individual is over $50K\$$. We want \textit{capital gain} and \textit{hours-per-week} to have an increasing causal relationship w.r.t. income. \textit{Race} and \textit{Sex} should have no causal effect. Finally, we assume that a person earns the most when they are about 50 years old, and hence hold this as the baseline intervention. Results shown in Fig \ref{fig:adult} and Table \ref{tab:adult_titanic} were obtained similar to the discussion in AutoMPG and Boston Housing dataset of main paper and once again support our claims.
\begin{table}
\footnotesize
\centering
\scalebox{0.8}{
\begin{tabular}{lcccccc}
\toprule
 Feature & \multicolumn{2}{c}{RMSE} & \multicolumn{2}{c}{Frechet Score} & \multicolumn{2}{c}{Corr. Coeff.}\\
  \midrule
  &ERM&CREDO&ERM&CREDO&ERM&CREDO\\
\midrule
 \multicolumn{7}{c}{Adult-ERM test accuracy is 80.72\%, CREDO test accuracy is \textbf{81.2\%}}\\
 \midrule
Race&0.212&\textbf{0.006}&0.395&\textbf{0.009}&-&-\\
Gender&0.321&\textbf{0.005}&0.454&\textbf{0.007}&-&-\\
Capital-gain&3.908&\textbf{0.299}&9.298&\textbf{1.708}&\textbf{0.999}&0.995\\
Hours-per-week&0.579&\textbf{0.275}&1.938&\textbf{0.985}&0.861&\textbf{0.973}\\
Age&0.652&\textbf{0.227}&1.889&\textbf{1.631}&0.256&\textbf{0.982}\\
%Avg.&1.134&\textbf{0.162}&2.795&\textbf{0.868}&0.256&\textbf{0.973}\\
  \midrule
 \multicolumn{7}{c}{Titanic-ERM test accuracy is 80.99\%, CREDO test accuracy is \textbf{81.30}\%}\\
 \midrule
 Age&0.17&\textbf{0.06}&0.30&\textbf{0.08}&-0.40&\textbf{0.99}\\
\bottomrule
\end{tabular}
}
\caption{ Results on ADULT, Titanic datasets}
\label{tab:adult_titanic}
\vspace{-7pt}
\end{table}
\begin{figure}
    \centering
    \includegraphics[width=.4\textwidth]{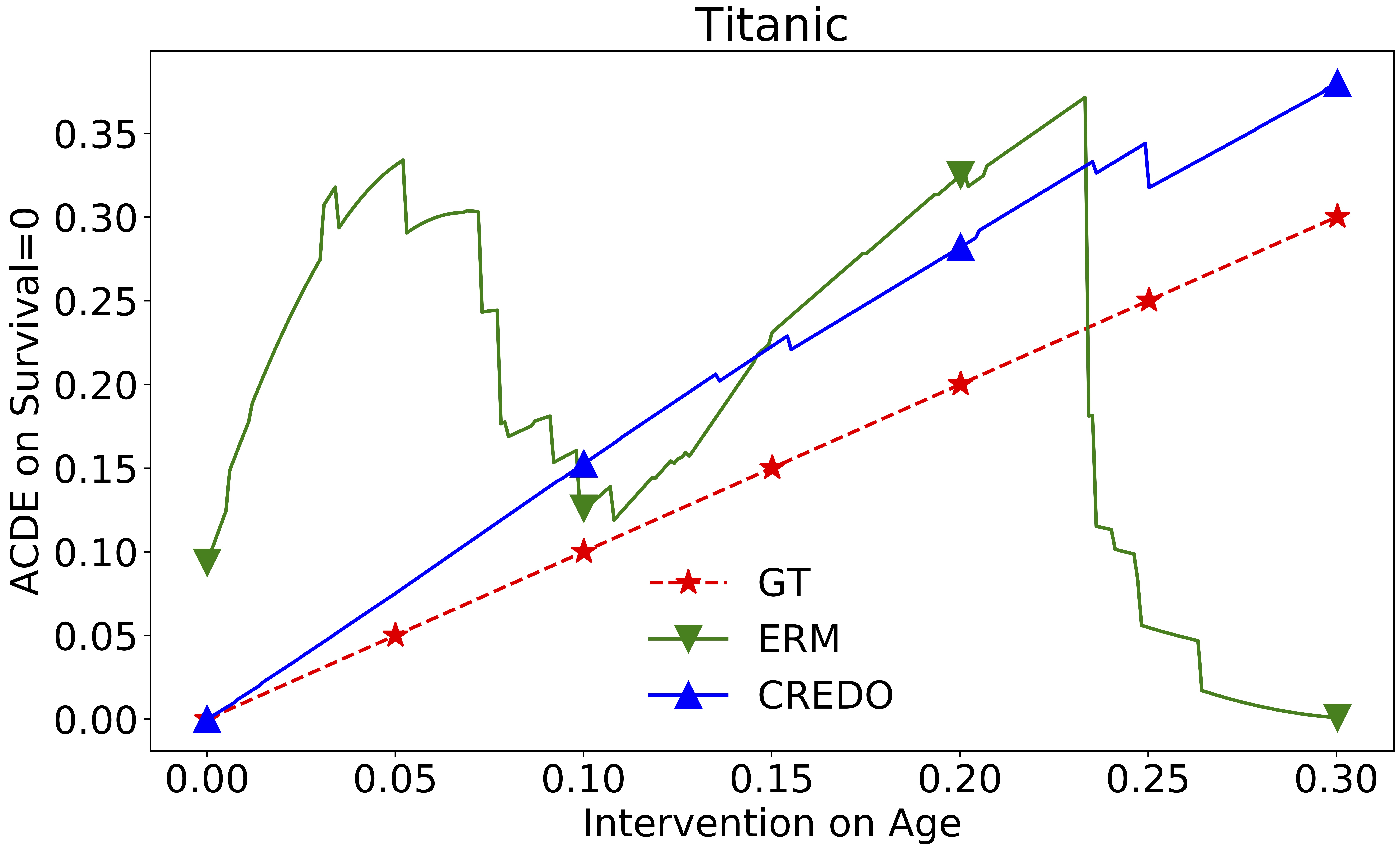}
    \caption{\small Enforcing Monotonicity on age: Comparison of ACE plots of ERM, CREDO on Titanic dataset}
    \label{fig:titanic}
\end{figure}

\vspace{-3pt} \noindent \textbf{Titanic:}
In the Titanic dataset~\cite{Dua:2019}, we predict the survival probability of an individual given a set of features. 
If we assume that any evacuation protocol gives more preference to children, we need the ACE of \textit{age} on \textit{survival = 1} class to be high when age value is small and ACE of age on \textit{survival = 0} class to be low when age value is small. We make no assumption if age value is greater than some threshold (0.3 for this experiment) and regularize the model for inputs that have age value less than 0.3. From the results (Fig \ref{fig:titanic}, Table \ref{tab:adult_titanic}), it is evident that CREDO is able to learn monotonic ACE of \textit{age} on \textit{survival}.

\vspace{-3pt}
\subsection{Comparison with PWL and DLN}
Table~\ref{tab:synthetic_tabular} below shows the quantitative results corresponding to the synthetic tabular dataset results in main paper.
\begin{table}
    \centering
    \footnotesize
    \begin{tabular}{lcccc}
    \toprule
     & RMSE & Frechet Score & Corr. Coeff. &Test Loss \\
    \midrule
     \multicolumn{5}{c}{Synthetic Tabular-1 ($z=\log(1+2x), x\in[0,1]$)}\\
    \midrule
    ERM&0.10&1.7e-3&0.99&\textbf{1e-2}\\
    PWL&0.10&1.7e-3&0.99&\textbf{1e-2}\\
    DLN&0.27&0.31&0.91&\textbf{1e-3}\\
    CREDO&\textbf{0.04}&\textbf{1.6e-3}&\textbf{1.0}&\textbf{1e-2}\\
    \midrule
     \multicolumn{5}{c}{Synthetic Tabular-2 ($z=\sin(x)+e^y$, $x\in[0,1]$, $y\in[0,1]$)}\\
    \midrule
    ERM&0.10&$\mathbf{1e-3}$&0.98& $2.05e-1$\\
    PWL&0.10&$\mathbf{1e-3}$&0.98&$1.45e-1$\\
    DLN&0.25&0.26&0.90&$2.7e-1$\\
    CREDO&\textbf{0.01}&$\mathbf{1e-3}$&\textbf{0.99}&$\mathbf{1.04e-1}$\\
    \bottomrule
    \end{tabular}
    \caption{ Comparison of ERM, CREDO, PWL, DLN on Synthetic Tabular-1, Synthetic Tabular-2 datasets}
    \label{tab:synthetic_tabular}
\end{table}
\subsection{Known Causal Graph: BNLearn Datasets}
\noindent \textbf{MEHRA:}
\begin{table}
    \footnotesize
    \centering
    \scalebox{0.85}{
    \centering
    \begin{tabular}{lcccccc}
    \toprule
     Feature & \multicolumn{2}{c}{RMSE} & \multicolumn{2}{c}{Frechet}& \multicolumn{2}{c}{Corr}\\
     \midrule
    &ERM&CREDO&ERM&CREDO&ERM&CREDO\\
     \midrule
    \multicolumn{7}{c}{\textbf{MEHRA}}\\
    \midrule
    \multicolumn{7}{c}{ACDE ($\lambda_1=2.2$) -(Test Acc:ERM: 80.75\%, CREDO: \textbf{82.00\%)}}\\
      \midrule
     Latitude&0.309& \textbf{0.045}& 0.366& \textbf{0.449}& 0.115 &\textbf{0.997}\\
     O3&0.588& \textbf{0.011} &0.062 &\textbf{0.991}& -0.494 &\textbf{1.000}\\
    SO2&0.790&\textbf{ 0.196}& \textbf{3.057} &1.840& 0.983 &\textbf{0.995}\\
    % Avg. &0.562 &\textbf{0.084}& \textbf{1.162}& 1.093& -0.494&\textbf{0.995}\\
    \midrule
        \multicolumn{7}{c}{ANDE ($\lambda_1=2.1$) - (Test Acc:ERM: 80.75\%, CREDO: 80.05\%)}\\
    \midrule
    Latitude&\textbf{0.309}& 0.400& 0.367& \textbf{0.481}& \textbf{0.116} &0.000\\
     O3&0.588& \textbf{0.023}& 0.063& \textbf{1.038}& -0.497& \textbf{1.000}\\
     SO2& 0.790& \textbf{0.045}& \textbf{3.054} &1.519 &0.983 &\textbf{1.000}\\
    % Avg. &0.562& \textbf{0.156} &\textbf{1.161} &1.013& -0.497& \textbf{0.000}\\
     \midrule
        \multicolumn{7}{c}{ATCE ($\lambda_1=1.5$) - (Test Acc:ERM: 80.75\%, CREDO: 79.30\%)}\\
    \midrule
    Latitude&0.335& \textbf{0.021}& 0.461& \textbf{0.505}& 0.090 &\textbf{0.999}\\
     O3&0.592 &\textbf{0.023}& 0.063& \textbf{1.004}& -0.740 &\textbf{1.000}\\
     SO2&0.823& \textbf{0.033}& \textbf{3.110} &1.456& 0.984& \textbf{1.000}\\
    % Avg. &0.583& \textbf{0.026}& \textbf{1.211} &0.989 &-0.740 &\textbf{0.999}\\
    \bottomrule
    \end{tabular}
    }
    \caption{ Results on MEHRA}
    \label{tab:sangiovese_mehra_appendix}
\end{table}
We generate data from the conditional linear Gaussian Bayesian network that models Multidimensional Environment-Health Risk Analysis (MEHRA) (Fig.~\ref{fig:mehra_dag}). We convert the output, \textit{total precipitation (TP)}, into a categorical variable and train a NN that predicts if it is greater than average. We choose three priors: a nonlinear (inverted-V shaped) ACE of \textit{Latitude} on \textit{TP}; and a linearly increasing ACE of \textit{O3}, \textit{SO2} on \textit{TP}.
Results in Table \ref{tab:sangiovese_mehra_appendix} show that CREDO helps conform to the causal prior in this case too.

\begin{figure}%[15]{r}{0.6\linewidth}
\centering
\includegraphics[width=0.47\textwidth]{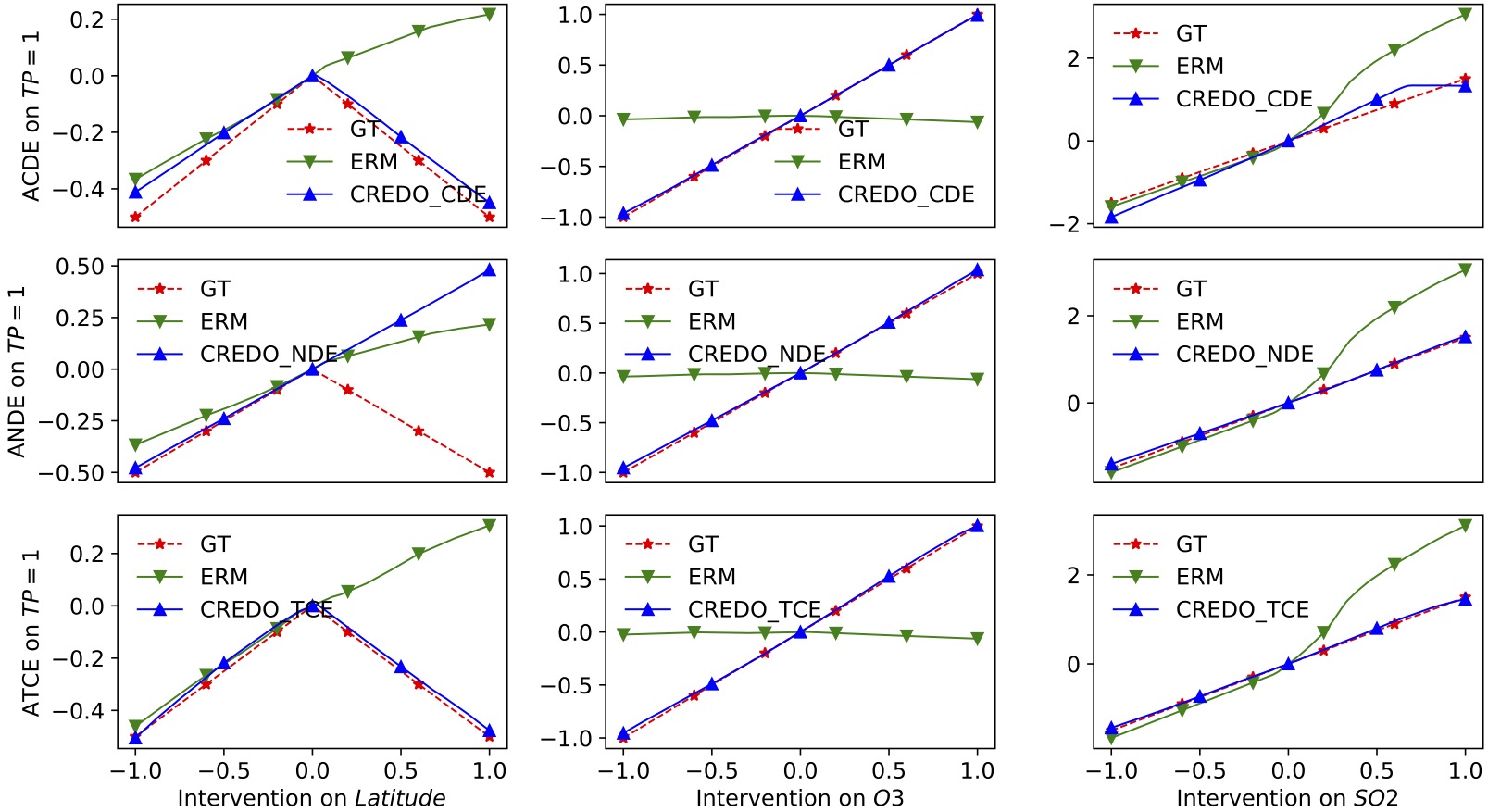}
\caption{Results on MEHRA: Comparison of ACDE, ANDE, ATCE learned by ERM and CREDO models.}
\label{fig:mehra}
\end{figure}

\vspace{-3pt} \noindent \textbf{Asia/Lung Cancer:}
In this experiment, we regularize for tuberculosis \textit{(tub)} and lung cancer \textit{(lung)} to have monotonically increasing relationship on the outcome dyspnoea \textit{(dysp)}. Table \ref{tab:asia_monotonic} shows the results where the regularized model, without any change in accuracy, learns to incorporate prior knowledge. 
\begin{table}
\footnotesize
\centering
\scalebox{0.9}{
\begin{tabular}{lcccccc}
\toprule
 Feature & \multicolumn{2}{c}{RMSE} & \multicolumn{2}{c}{Frechet Score} & \multicolumn{2}{c}{Corr. Coeff.}\\
  \midrule
 \multicolumn{7}{c}{ERM test accuracy is 85.20\%, CREDO test accuracy is 85.20\%}\\
 \midrule
&ERM&CREDO&ERM&CREDO&ERM&CREDO\\
\midrule
Tub&0.387&\textbf{0.279}&0.729&\textbf{0.579}&0.874&\textbf{0.983}\\
Lung&0.858&\textbf{0.265}&1.55&\textbf{0.620}&-0.989&\textbf{0.964}\\
%Avg.&0.622&\textbf{0.272}&1.139&\textbf{0.599}&-0.057&\textbf{0.973}\\
\bottomrule
\end{tabular}
}
\caption{ Enforcing Monotonic Effects: Results on ASIA dataset}
\label{tab:asia_monotonic}
\vspace{-7pt}
\end{table}

\vspace{-3pt} \noindent \textbf{SACHS:}
In the SACHS dataset (generated using underlying causal graph shown in Figure \ref{fig:sachs}), without going into the semantic meanings of the features, it is evident that \textit{Jnk, PIP2, PIP3, Plcg, P38} are non-causal predictors of \textit{Akt}. With CREDO, we get good accuracy while maintaining the zero causal effect of the non-causal predictors on the outcome (Table ~\ref{tab:sachs_zero_dce}). 
\begin{table}
    \centering
    \footnotesize
    \begin{tabular}{lcccc}
    \toprule
     Feature & \multicolumn{2}{c}{RMSE} & \multicolumn{2}{c}{Frechet Score}\\
     \midrule
     \multicolumn{5}{c}{ERM test accur: 82.05\%, CREDO test accur: 85.10\%}\\
     \midrule
    &ERM&CREDO&ERM&CREDO\\
    \midrule
    Jnk&0.030&\textbf{0.000}&0.038&\textbf{0.000}\\
    P38&0.026&\textbf{0.000}&0.036&\textbf{0.000}\\
    PIP2&0.073&\textbf{0.000}&0.199&\textbf{0.000}\\
    PIP3&0.103&\textbf{0.001}&0.149&\textbf{0.001}\\
    Plcg&0.020&\textbf{0.000}&0.035&\textbf{0.000}\\
    %Avg.&0.050&\textbf{0.000}&0.100&\textbf{0.000}\\
    \bottomrule
    \end{tabular}
    \caption{ Enforcing Zero Causal Effects: Results on SACHS}
    \label{tab:sachs_zero_dce}
\end{table}
\section{Ablation Studies and Analysis}
\label{sec_ablations}
We now report our results from ablation studies that study various aspects of CREDO, such as how CREDO behaves under different priors for the same problem, time complexity analysis, etc.

\begin{figure}
    \centering
    \includegraphics[width=0.45\textwidth]{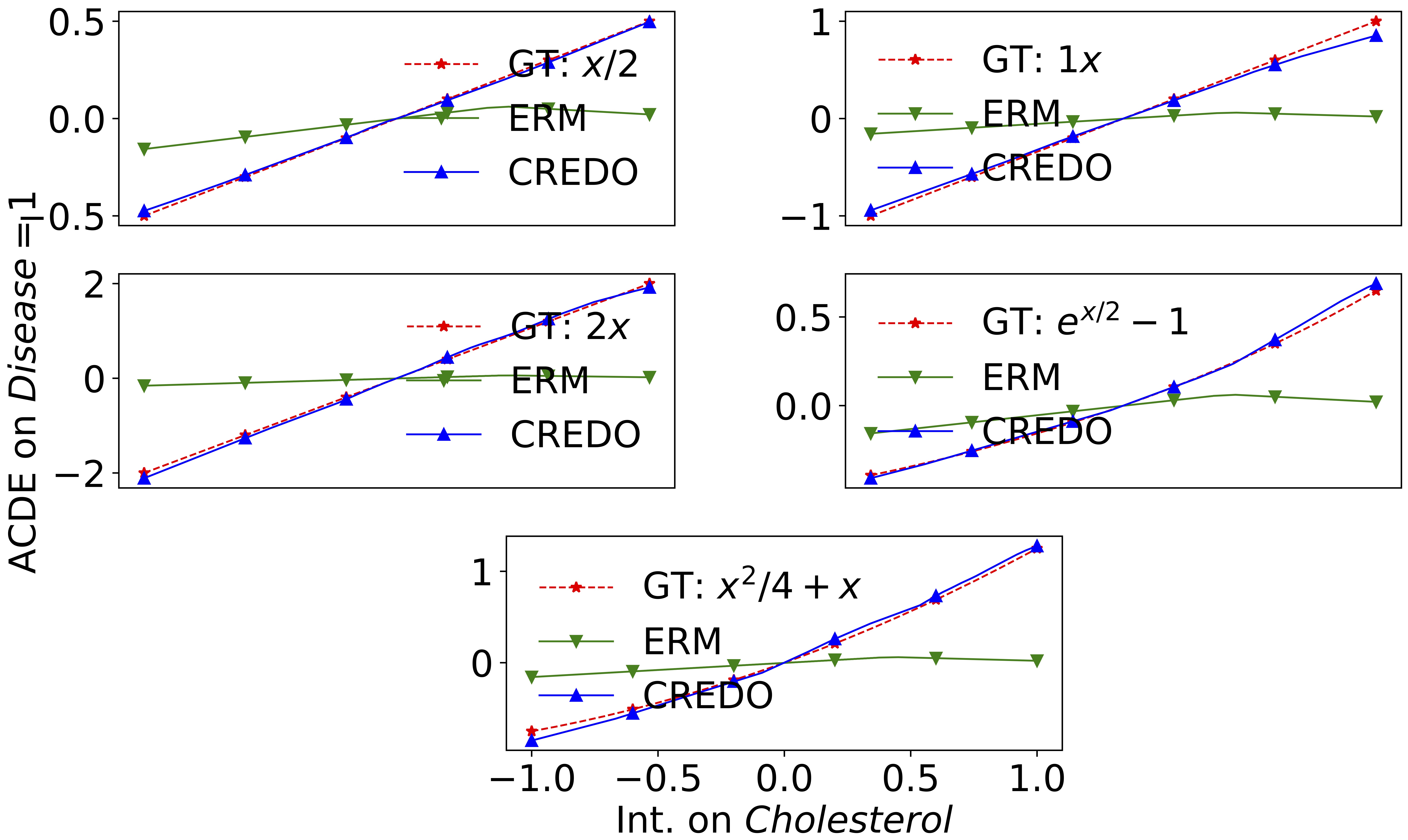}
    \caption{Plots of ACE Cholesterol(chol) on heart disease for various priors.}
    \label{fig:multiple_priors}
\end{figure}

\vspace{-3pt} \noindent \textbf{Enforcing Monotonicity with Different Priors:}
Using heart disease dataset~\cite{Dua:2019}, we show how our method can be used when it is known that the prior shape is monotonic, but the exact slope of the monotonic prior is not known. In particular, we show that we are able to match any assumed shape of the monotonic prior in this setting. (This raises questions on how a given prior can be validated for correctness, which is an interesting direction of future work by itself.) Fig \ref{fig:multiple_priors} shows the results, where CREDO is able to match any of the different priors. In such scenarios, one can choose a parametrization of the prior that maximizes generalization performance, while respecting domain knowledge.

\vspace{-3pt} \noindent \textbf{Time Complexity:} 
All experiments were conducted on one NVIDIA GeForce 1080Ti GPU. We compared the training time of ERM and CREDO. On the Boston Housing dataset, ERM takes 36.02secs while CREDO takes 40.40secs to train for 100 epochs. On AutoMPG, ERM takes 16.39secs while CREDO takes 17.70secs to train for 50 epochs. CREDO trains in almost the same time as ERM with a marginal increase, while providing the benefit of causal regularization.

\vspace{-3pt} \noindent \textbf{\textbf{Effect of Choice of $\lambda_1$, Regularization Coefficient:}} 
\begin{figure}%{r}{0.6\linewidth}
\centering
    \includegraphics[width=0.45\textwidth]{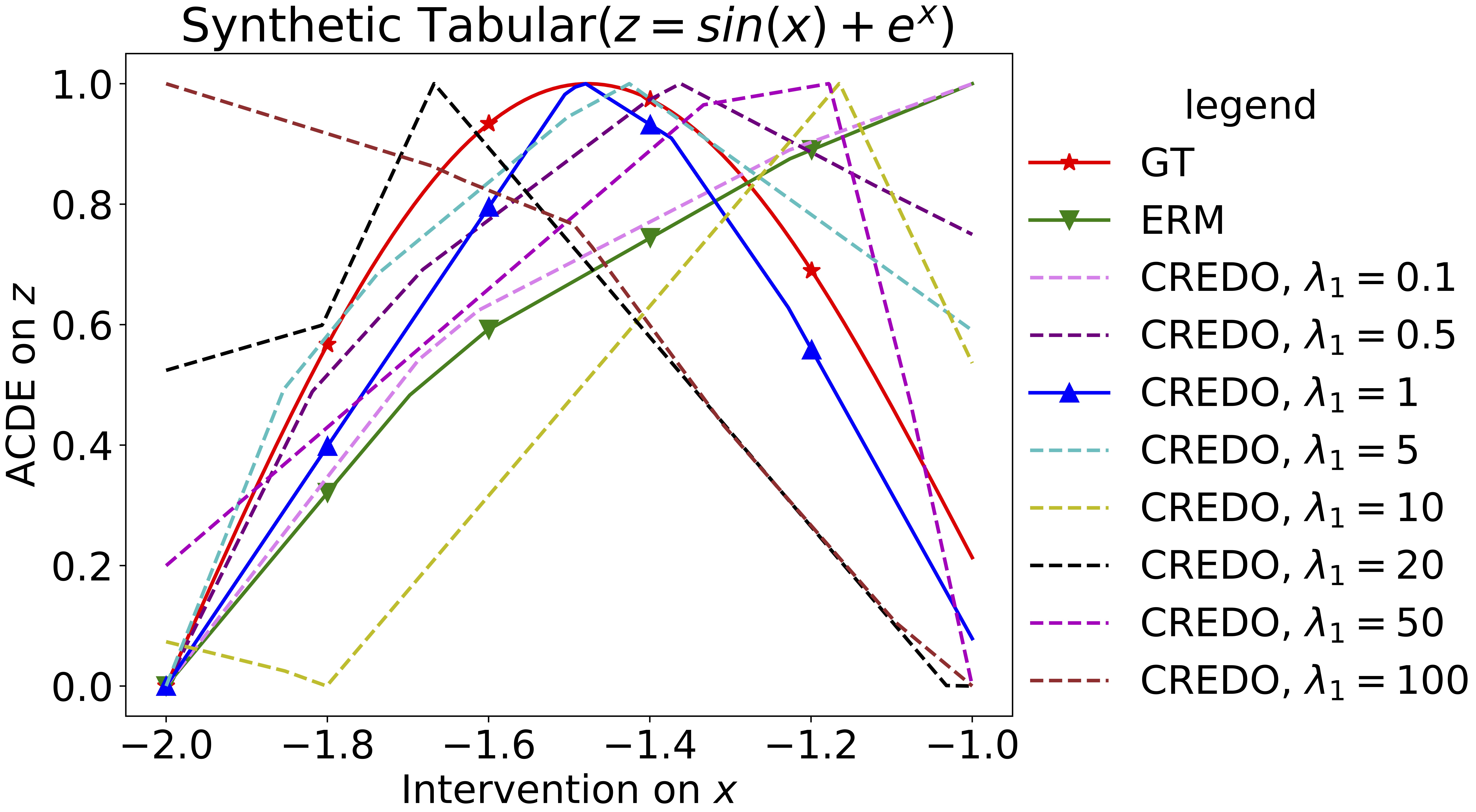}
    \caption{ $ACDE$ of $x$ on $z=\sin(x)+e^y$ learned by models}
    \label{fig:arbitrary}
\end{figure}
A grid search is performed to fix the regularization coefficient $\lambda_1$. We report the specific values chosen for each dataset in the corresponding result. To study the effect of regularization coeff $\lambda_1$, we again consider the function $f(x,y)=\sin(x)+e^y$, now with inputs $x\in[-2,-1], y\in[-2,-1]$ (Synthetic Tabular 3) (note the change in interval, this allows the domain prior to be an arbitrary shape close to $\sin(x)$ rather than monotonic). Fig \ref{fig:arbitrary} shows the plots of $ACDE$ of $x$ learned by the model with and without CREDO with different co-efficients. We notice that while a specific value ($\lambda_1=1$) provides the best match with the GT, most other choices for $\lambda_1$ do better than ERM in matching the prior.

\vspace{-3pt} \noindent \textbf{Effect of Incorrect Prior:}
\begin{table}%{r}{0.6\linewidth}
    \centering
    \footnotesize
    \begin{tabular}{lccc}
    \toprule
         Method & Prior Slope of \textit{Race}&Test Acc\\
         \midrule
         ERM & -&86.27\\
        CREDO & 0  & 86.04\\
        CREDO & -2  & 85.00\\
        CREDO & 2 & 83.50\\
        CREDO & 3 & 83.40\\
         \bottomrule
    \end{tabular}
    \caption{ Effect of modulating Race prior on MEPS dataset}
    \label{tab:meps_ablation}
    \vspace{-7pt}
\end{table}
To understand the behavior of regularization with an incorrect prior on model performance, we study the MEPS dataset and observe that while making the model match incorrect priors reduces the model accuracy expectedly as presented in Table \ref{tab:meps_ablation}.

Our method may not work well when the provided priors are substantially different from the true causal relationships. This may happen in case of a mismatch between domain knowledge and creation of the causal graph, or a faulty understanding of the causal relationships. While we focused our efforts in this work with the assumption that the provided causal priors are true, studying the faulty nature of priors under/using our framework is an interesting direction of future work. 

\vspace{-3pt} \noindent \textbf{Correct Shape and Arbitrary Slope:}
In this experiment, we ask CREDO the question \textit{"If one knows that the causal prior is monotonic but not the exact slope, how well CREDO matches the results with true prior?"} To this end, we performed the following experiment. We create a synthetic tabular dataset (Synthetic Tabular 4) using the structural equations given below so that the true gradient of the causal effect of $X$ on $Y$ is known, which is 2. 
\begin{equation*}
\begin{aligned}
    W :=& \mathcal{N}_w(0, 1)\\
    Z :=& -2W+\mathcal{N}_z(4,1)\\
    X :=& 0.5Z+\mathcal{N}_x(2,1)\\
    Y :=& 2X + Z + W +\mathcal{N}_x(0,0.1)
\end{aligned}
\end{equation*}

In the dataset generated using these equations, we use $X,Z,W$ as inputs and $Y$ as output to a neural network, and the input given to CREDO is that the slope is linearly monotonic. We provide different assumed domain priors (slope values) to CREDO, to simulate a setting where the incorrect slope value is provided as input. We make two observations:  (i) As we get closer to the true slope, the CREDO classifier's accuracy improves (Table~\ref{tab:synthetic_slope_vs_accuracy}). The highest accuracy is for the assumed slope=2, which is the true slope. Note that our method here has no information about the true slope. (ii) Assuming that only the linear monotonicity property of the gradient was input to CREDO, if we use this simple linear search to find the gradient hyperparameter $\alpha$, our method would return the correct assumed gradient prior=2 since that achieves the highest classification accuracy.  With these results, it is evident that the closer our assumed gradient is to the true gradient, the better the accuracy is.
\begin{table}
    \centering
    \footnotesize
    \begin{tabular}{cccc}
    \toprule
     \textbf{Method}& \textbf{Assumed Prior Slope} &\textbf{  Accuracy}\\
     \midrule
     ERM   & -   &87.00\%\\
     CREDO & 1   & 86.95\%\\
     CREDO & 1.2 & 87.00\%\\  
     CREDO & 1.4 & 86.70\%\\
     CREDO & 1.6 & 86.35\%\\
     CREDO & 1.8 & 87.20\%\\
     \textbf{CREDO} & \textbf{2.0} & \textbf{87.60}\%\\
     CREDO & 2.2 & 87.30\%\\
     CREDO & 2.4 & 87.05\% \\
     CREDO & 2.6 & 86.85\% \\
     CREDO & 2.8 & 86.35\% \\
     CREDO & 3.0 & 86.60\% \\
     
    \bottomrule
    \end{tabular}
    \caption{Synthetic Tabular 4: Slope vs Accuray}
    \label{tab:synthetic_slope_vs_accuracy}
\end{table}

\begin{figure}
	\centering
    \includegraphics[width=0.5\textwidth]{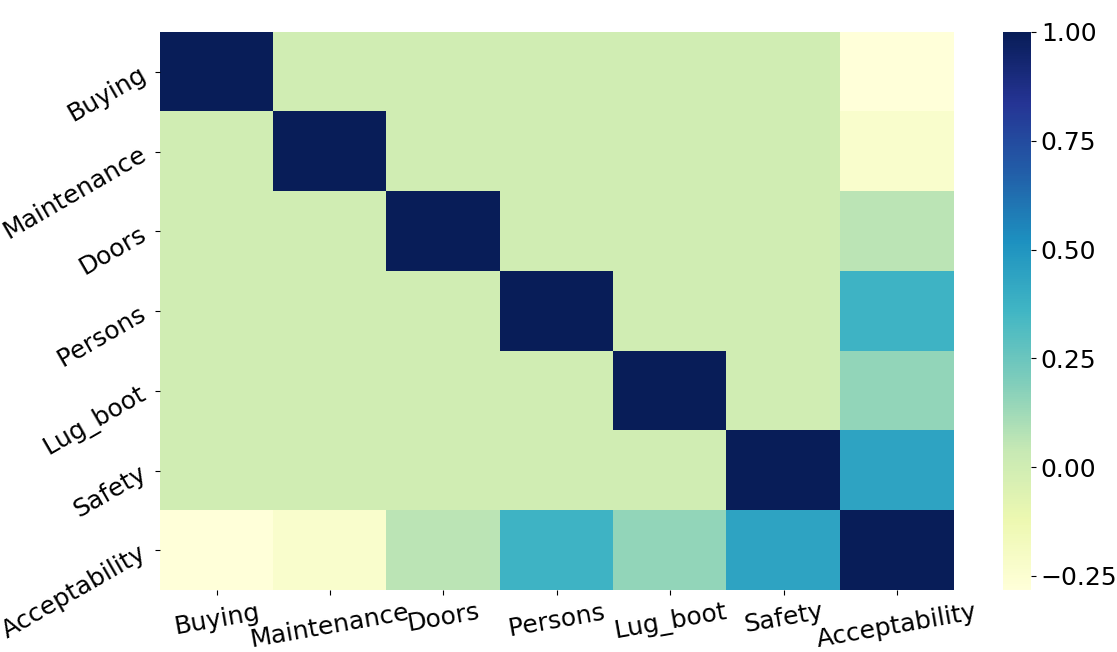}
    \caption{Data Correlation in Car Evaluation Dataset.}
    \label{fig:cars_corr}
\end{figure}

\vspace{-3pt} \noindent \textbf{Is CREDO always useful?}
\begin{figure}
    \centering
    \includegraphics[width=0.4\textwidth]{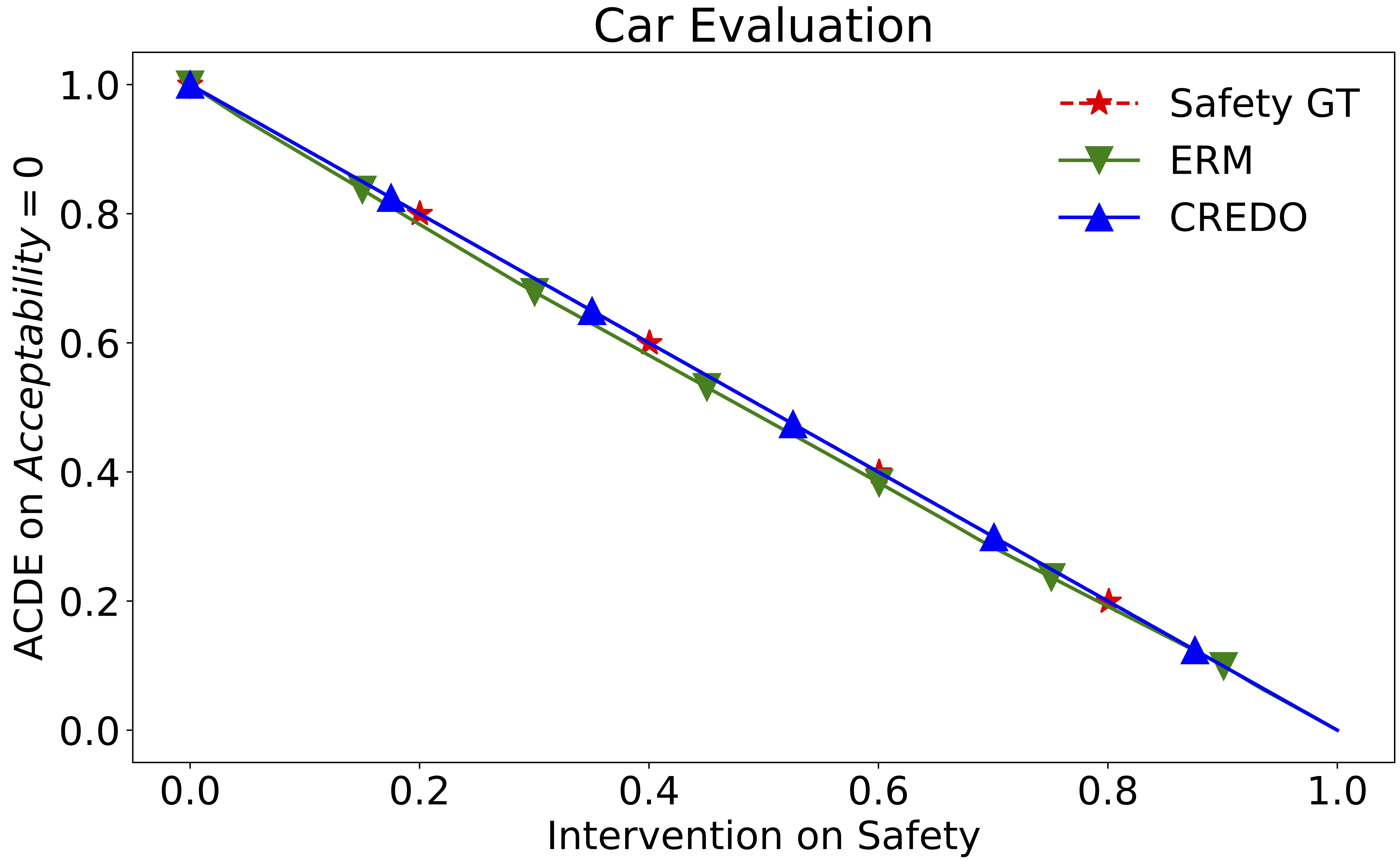}
    \caption{Comparison of ACE plots of ERM, CREDO.}
    \label{fig:cars}
\vspace{-7pt}
\end{figure}
When the domain prior also matches the most significant correlations in a dataset, ERM can perform well by itself. We note however that this is not common especially in most real-world datasets. As an example, in the Car evaluation dataset from the UCI ML repository, the task is to predict the \textit{acceptability} of a car given a set of features. Intuitively, \textit{safety} levels of a car should have monotonic causal effect on its acceptability. Fig \ref{fig:cars} and Table \ref{tab:cars} show the results of running ERM and CREDO on this dataset. On closer analysis of the dataset (Fig \ref{fig:cars_corr}), we observe that among all features in the dataset: \textit{Buying,Maintenance,Doors,Persons,Lug\_boot,Safety}, high correlation is observed between \textit{Safety} and \textit{acceptability}. ERM captures this and performs comparably to CREDO under such a scenario.

\begin{table}
\footnotesize
\begin{center}
\begin{tabular}{lccccc}
\toprule
 &RMSE &Frechet Score&Corr. Coeff.&Test Loss \\
\midrule
ERM& 0.03&0.0011&0.99& \textbf{99.07} \\
CREDO& \textbf{0.003} &\textbf{0.0010}&\textbf{0.99}& 97.86 \\
\bottomrule
\end{tabular}
\caption{ Enforcing Monotonic Causal Effect: Results on Car Evaluation dataset}
\label{tab:cars}
\vspace{-7pt}
\end{center}
\end{table}
\vspace{-7pt}
\section{ACE Algorithm}
\label{acealgo}
For all our experiments, qualitative results are obtained using ACE plots. We use the algorithm proposed in \cite{chattopadhyay2019neural} for computation of ACE. For completeness, we briefly present the ACE calculation in Algorithm \ref{algo:ace}. The algorithm summarizes the method to find $E(\hat{Y}_t)$; it is easy to find ACE subsequently as $ACE^{\hat{Y}}_T = E(\hat{Y}_t)-E(\hat{Y}_{t^*})$. For more details please refer to \cite{chattopadhyay2019neural}. This algorithm depends on a Taylor's series expansion of neural network output, and hence the use of first-order and second-order gradients in the method. 
\begin{algorithm}
\footnotesize
\SetAlgoLined
\KwResult{$E(\hat{Y}_{t})$ for each $t=\alpha$}
\textbf{Inputs:} $f, t$, $t$'s range: $[low, high]$, number of interventitons: $n$, $data\ mean:\ \mu, data\ covariance\ matrix:\ cov$

\textbf{Initialize:} $cov[t][:]\coloneqq 0, cov[:][t]\coloneqq 0$,$\alpha=low$, $IE\coloneqq[]$
 
\While{$\alpha \leq high$}{
$\mu[i]=\alpha$\;

$IE.append(f(\mu)+\frac{1}{2}trace(matmul(\nabla^2f(\mu), cov)))$ \;

$\alpha = \alpha +\frac{high-low}{n}$\;
}
return $IE$\;
\caption{ACE learned by the neural network}
\label{algo:ace}
\end{algorithm}
\vspace{-7pt}
\section{BNLearn Datasets: Causal Graphs}
\label{dags}
\vspace{-6pt}
\begin{figure}
\centering
\includegraphics[width=.5\textwidth]{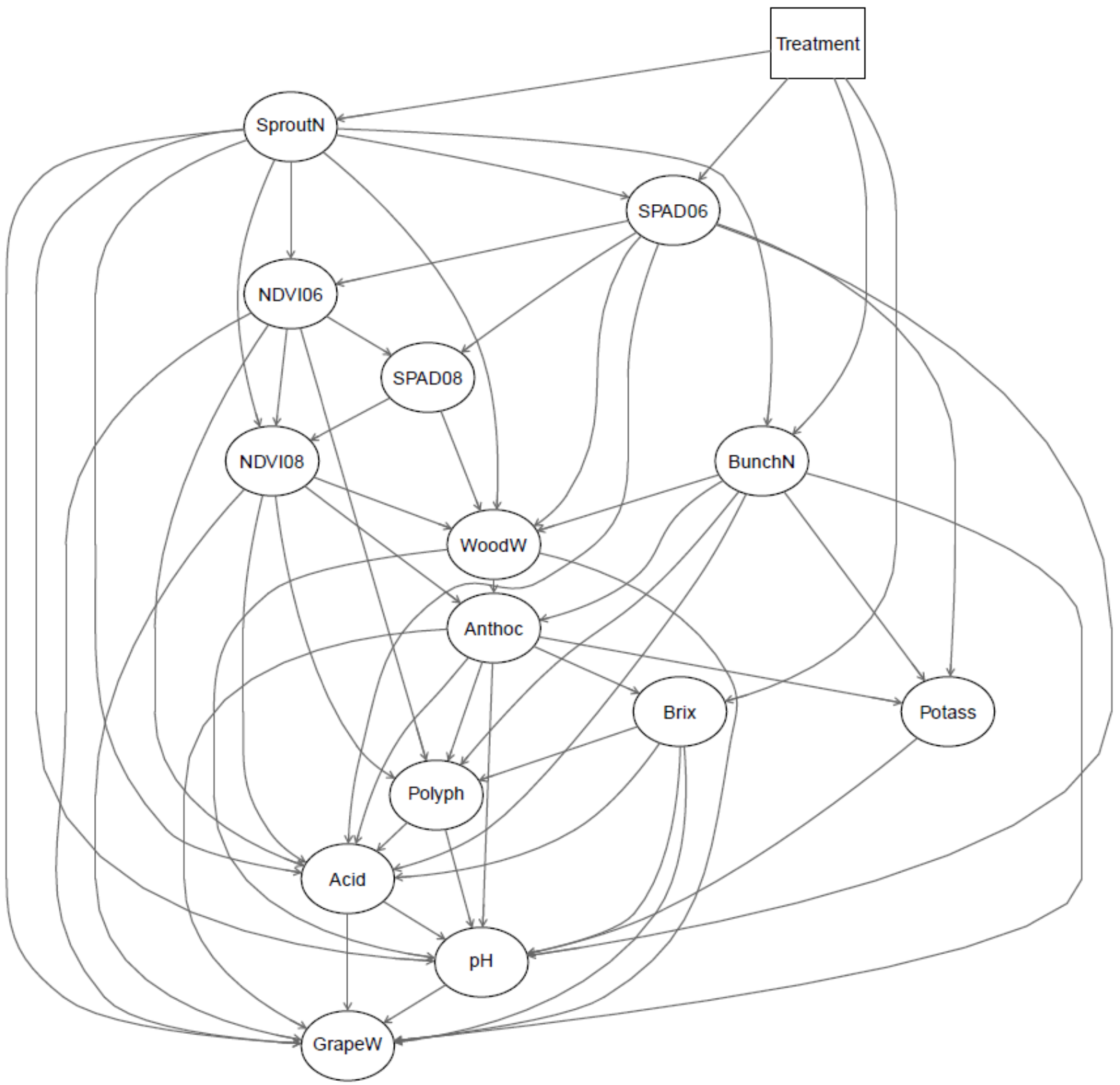}
\caption{Causal Bayesian network of SANGIOVESE dataset}
\label{fig:sangiovese}
\vspace{-12pt}
\end{figure}
\begin{figure}
\centering
\includegraphics[width=.47\textwidth]{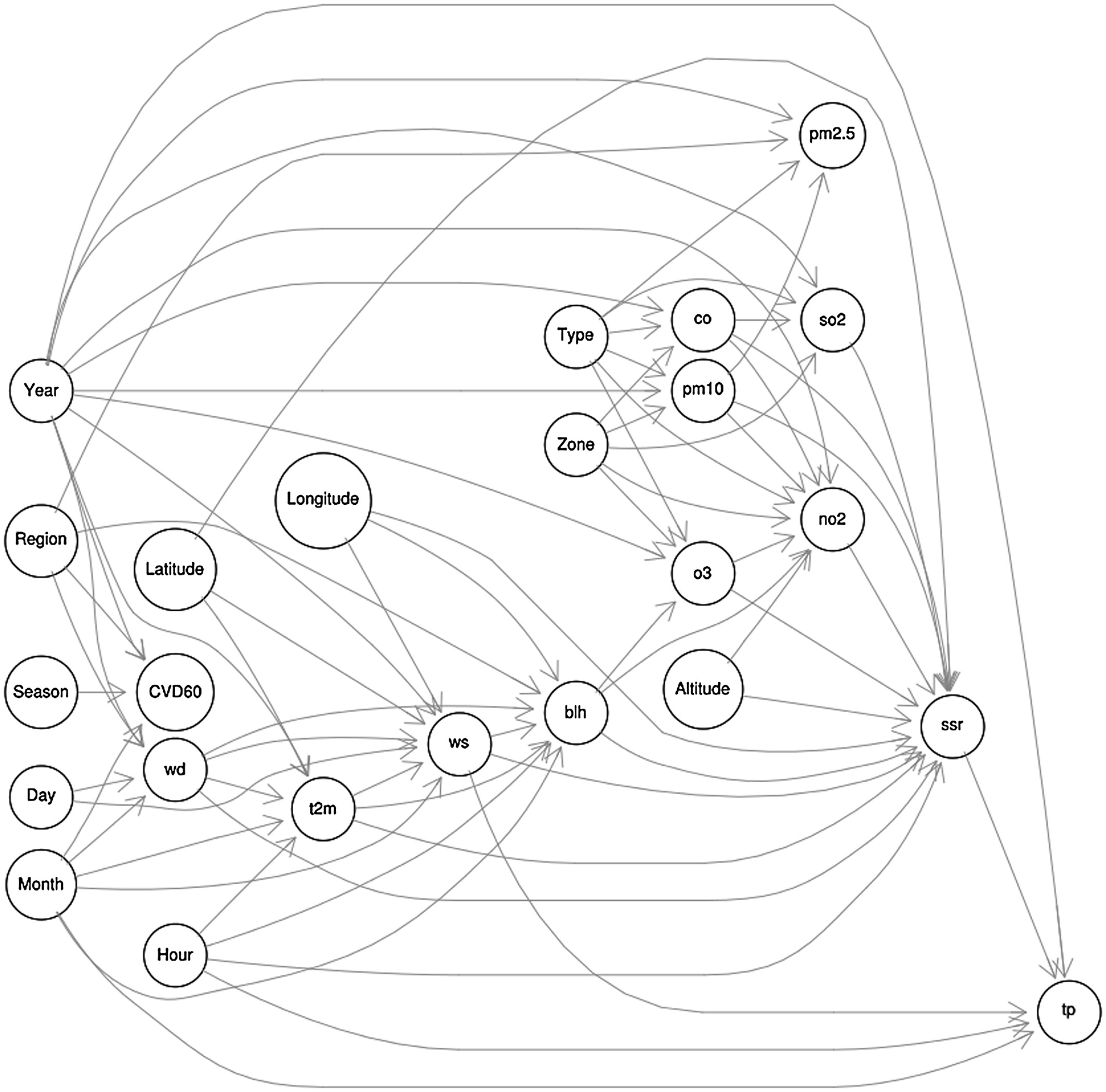}
\caption{Causal Bayesian network of MEHRA dataset}
\label{fig:mehra_dag}
\vspace{-12pt}
\end{figure}
\begin{figure}
\centering
\includegraphics[width=.35\textwidth]{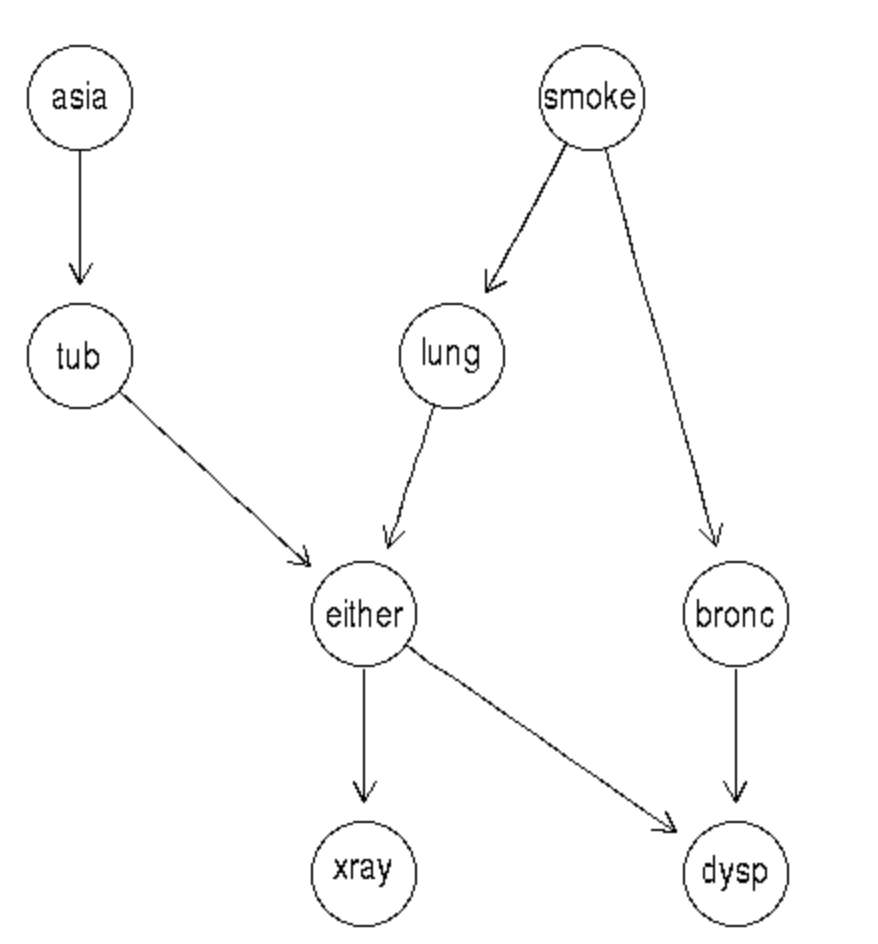}
\caption{Causal Bayesian network of ASIA/Lung Cancer dataset}
\label{fig:asia}
\end{figure}
\begin{figure}
    \centering
    \includegraphics[width=.38\textwidth]{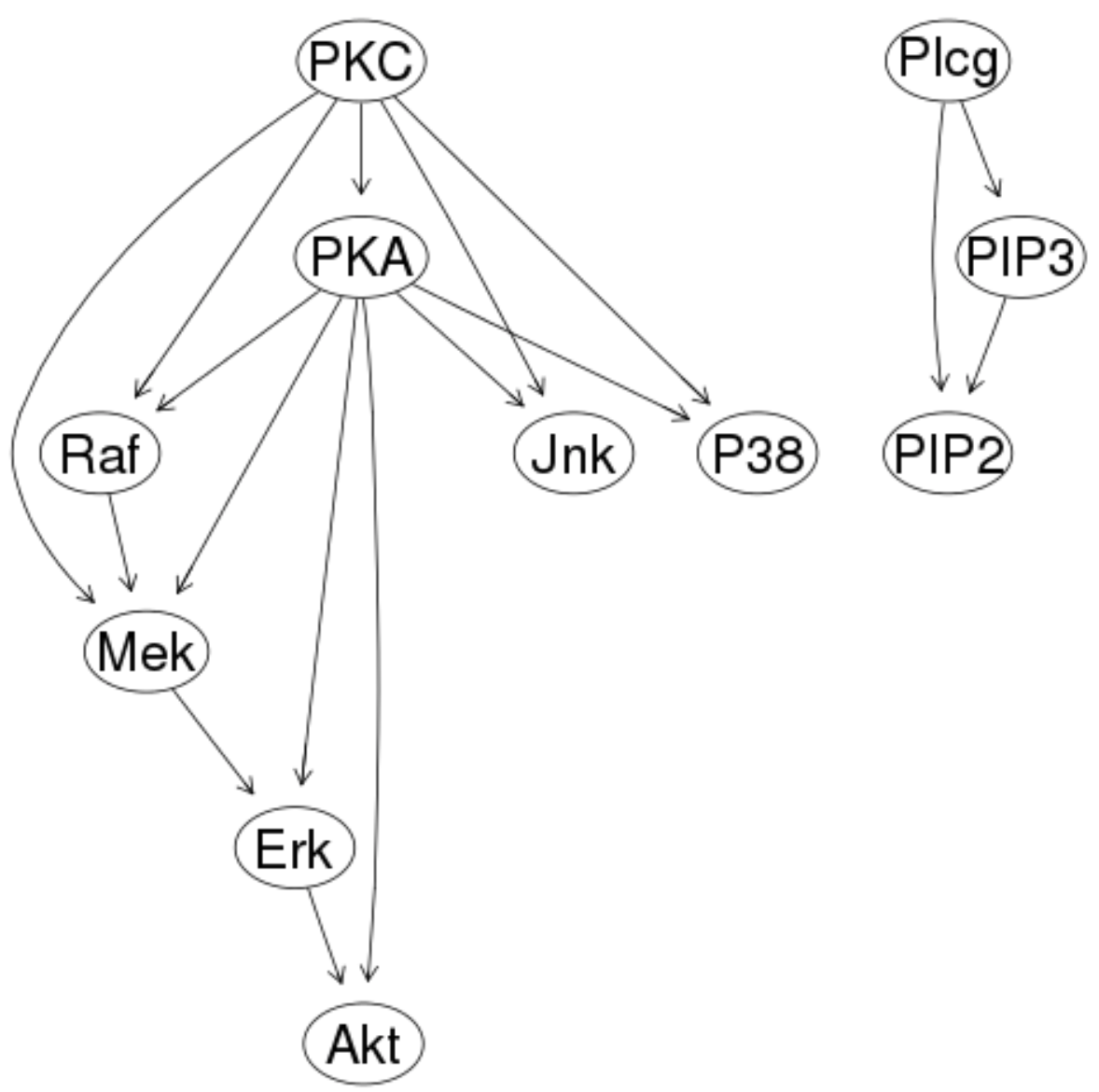}
    \caption{Causal Bayesian network that generates SACHS dataset}
    \label{fig:sachs}
    \vspace{-12pt}
\end{figure}

The causal DAGs of SANGIOVESE, MEHRA, Asia, and Sachs datasets from the BNLearn repository \cite{scutari2014bayesian} are shown in Figs \ref{fig:sangiovese}, \ref{fig:mehra_dag}, \ref{fig:asia}, and \ref{fig:sachs} respectively.
\vspace{-7pt}
\section{Architectural/Training Details}
\label{arch}
\vspace{-6pt}
We use a multi-layer perceptron with ReLU non-linearity across our experiments, each trained using ADAM optimizer with Dropout and $L_2$ weight decay. Table \ref{tab:training_details} shows the details of neural network architectures and training details of our models for various datasets. $80\%$ of the dataset is used for training and remaining $20\%$ for testing. We observed that gradients on output logits work better than softmax or logsoftmax activated outputs. 
\begin{table*}[h]
    \centering
    \footnotesize
    \begin{tabular}{lcccccccc}
    \toprule
         S.No.&Dataset &Dataset& Input Size,&Learning Rate & Batch Size & $\lambda_1$&Number of &Size of  \\
         &&Size&Output Size&&&(ACDE)&Layers&Each Layer\\
         \midrule
         1&COMPAS&6,172&11,2&1e-2 &128&50&4&16,16,16,16\\
         \midrule
         2&MEPS& 15,830&138,2&1e-3&64&2&3&128,256,256\\
         \midrule
         3&Law School&20,797&15,2&1e-3&64&2&3&64,32,64\\
         \midrule
         4&AutoMPG&398&7,2&1e-2&32&1.5&4&16,16,16,16\\
         \midrule
         5&Boston Housing&506&13,2 &1e-2&32&1&4&16,16,16,16\\
         \midrule
         6&Titanic&1,309&10,2&1e-3&64&3&2&64,128\\
         \midrule
         7&Car Evaluation&1,728&6,2&1e-3&64&2&3&64,128,128\\
         \midrule
         8&Heart Disease&303&13,2&1e-2&64&1&3&16,16,16\\
         \midrule
         9&Adult&48842&14,2&1e-2&1024&0.2&3&64,64,64\\
         \midrule
         10&SANGIOVESE&10,000&29,2&1e-2&256&2.3&3&16,16,16\\
         \midrule
         11&SACHS&10,000&10,3 &1e-3&64&10&2&16,32\\
         \midrule
         12&ASIA&10,000&7,2 &1e-3&64&1&2&32,64\\
         \midrule
         13&MEHRA&10,000&152,2&1e-3&64&2.2&3&32,32,32\\
         \midrule
         14&Synthetic Tabular 1\\
         &$z=\log(1+2x)$&1,000&1,1 &1e-2&64&10&2&4,8\\
         &$x\in[0,1]$\\
         \midrule
         15&Synthetic Tabular 2\\
         &$z=\sin{x}+e^y$&1,000&2,1&1e-2&64&0.5&2&8,16\\
         &$x,y\in[0,1]$\\
         \midrule
         16&Synthetic Tabular 3\\
         &$z=\sin{x}+e^y$&1,000&2,1&1e-3&64&20&2&8,16\\
         &$x,y\in[-2,-1]$\\
         \midrule
         17&Synthetic Tabular 4&10,000&3,2&1e-2&64&1.0&3&12,12,12\\
         &$W := \mathcal{N}_w(0, 1)$\\
         &$\vdots$\\
         \bottomrule
    \end{tabular}
    \caption{Architectural and training details of all datasets.}
    \label{tab:training_details}
\end{table*}
\end{document}